\pdfoutput=1
\documentclass[lettersize,journal]{IEEEtran}
\usepackage{amsmath,amsfonts,amssymb}
\usepackage{cite}
\usepackage{algorithmic}
\usepackage{array}
\usepackage[caption=false,font=normalsize,labelfont=sf,textfont=sf]{subfig}
\usepackage{textcomp}
\usepackage{stfloats}
\usepackage{url}
\usepackage{verbatim}
\usepackage{color}
\usepackage{graphicx}
\graphicspath{{figures/}}
\def\BibTeX{{\rm B\kern-.05em{\sc i\kern-.025em b}\kern-.08em
    T\kern-.1667em\lower.7ex\hbox{E}\kern-.125emX}}
\usepackage{balance}
\usepackage{multirow}
\begin{document}
\title{Passive-Dynamic-Walking-Inspired Dynamics Guidance for Energy-Efficient Humanoid Locomotion}
\author{Hyeonjin~Choi,
        Joongheon~Kim,~\IEEEmembership{Senior Member, IEEE},
        and Daekyum~Kim,~\IEEEmembership{Member, IEEE}%
\thanks{This work has been submitted to the IEEE for possible publication. Copyright may be transferred without notice, after which this version may no longer be accessible.}%
\thanks{H. Choi and D. Kim are with the School of Mechanical Engineering, Korea University, Seoul 02841, Republic of Korea (e-mails: \{hyeonjin98,daekyum\}@korea.ac.kr). (Corresponding authors: Joongheon Kim; Daekyum Kim.)}%
\thanks{J. Kim is with the School of Electrical Engineering, Korea University, Seoul 02841, Republic of Korea (e-mail: joongheon@korea.ac.kr).}%
\thanks{D. Kim is also with the School of Smart Mobility, Korea University, Seoul 02841, Republic of Korea.}%
}

\markboth{}%
{Choi \MakeLowercase{\textit{et al.}}: PDW-Inspired Dynamics Guidance for Energy-Efficient Humanoid Locomotion}

\maketitle

\begin{abstract}
Learning energy-efficient humanoid locomotion requires discovering mechanically economical gait coordination, not merely reducing actuator effort. Reinforcement learning promotes efficiency through effort-related reward penalties, which guide the step-to-step mechanics of walking only indirectly. This article proposes a framework inspired by passive dynamic walking (PDW) that temporarily creates slope-equivalent conditions favorable to economical gait discovery and removes all PDW-specific guidance before nominal-dynamics optimization. During early training, a tilted-gravity field assists sagittal progression on flat collision geometry, complemented by curriculum-coupled reward terms. The core framework requires no reference trajectories, gait phases, or contact schedules. In a five-seed forward-locomotion study on a 29-DoF Unitree G1, the framework reduces mechanical cost of transport by 6.8--15.2\% over commanded speeds of 0.5--2.0~m/s without degrading velocity tracking. Mechanical-work decomposition attributes the reduction to positive actuator work, and reward-matched comparisons separate the guided regime's faster gait acquisition from the tilt's additional benefit to converged economy. The framework extends to unassisted omnidirectional locomotion, where its benefit persists once a walking-specific motion prior supplies kinematic coordination, the combination reducing speed-matched cost of transport by 18.7\%. On hardware, forward cost of transport falls by 16.3\% with the motion prior and by 4.5\% without it, the latter within the trial-to-trial spread.
\end{abstract}


\section{Introduction}

\IEEEPARstart{H}{uman} walking is known to be energy efficient, and explaining how this efficiency is achieved requires both mechanical and learning perspectives.
Mechanically, energy-efficient walking depends strongly on how mechanical work is organized from one step to the next, especially during the transition between stance legs \cite{DonelanEtAL,KuoEtAL2005}.
As a learning problem, this organization is not present from the outset: walking is not acquired in a single attempt, and biological learners gradually develop it through repeated practice, errors, and corrections \cite{AdolphEtAL2012,AdolphEtAL2003}.
In humans, this learning unfolds through interaction among the learner's body dynamics, the task, and the environment rather than independently of them \cite{Thelen1995}.
The two problems are therefore inseparable, since the mechanical organization that makes walking economical is precisely what has to be learned.

Such trial-and-error learning motivates reinforcement learning (RL) for humanoid locomotion.
Instead of manually designing a full sequence of robot motion specifications, such as desired joint motions, contact timings, and center-of-mass trajectories, RL enables the robot to improve its behavior through repeated trial-and-error interaction with the environment, guided by a reward function \cite{SuttonEtAL1998}.
Recent RL controllers have substantially advanced humanoid locomotion, enabling high-degree-of-freedom humanoid robots to track commanded velocities, recover from disturbances, traverse challenging terrain, and transfer policies from simulation to hardware \cite{RadosavovicEtAL2024,GuEtAL2024,LiEtAL2025a}.
The reward functions used in these systems commonly include task objectives such as tracking a commanded velocity, maintaining an upright and stable base, and avoiding undesirable contacts \cite{RudinEtAL,RadosavovicEtAL2024}.

\begin{figure}[!t]
\centering
\includegraphics[width=\columnwidth]{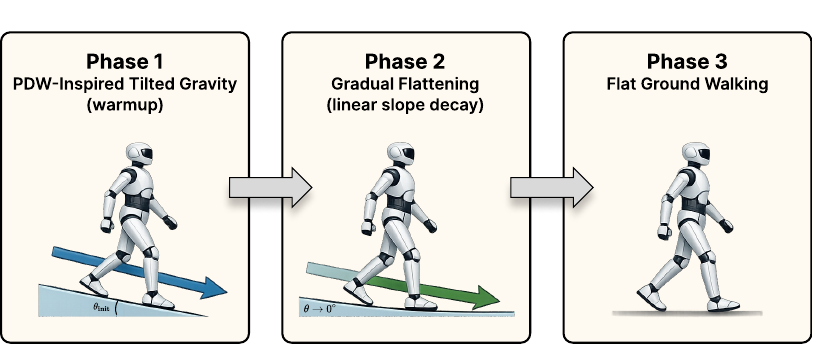}
\caption{Overview of the proposed PDW-inspired slope-to-flat curriculum. \emph{Phase 1}: early policy search is conducted under a PDW-inspired tilted-gravity field, which supplies gravitational assistance to fore--aft step-to-step progression. \emph{Phase 2}: the tilt decays linearly, returning the training dynamics toward nominal. \emph{Phase 3}: all PDW-specific guidance is fully withdrawn, and the policy must sustain the induced gait structure under powered flat-ground command tracking. The collision geometry remains flat throughout; the depicted incline illustrates only the energetic effect of the tilted-gravity field.}
\label{fig:overview}
\end{figure}

In RL, the mechanical organization required for efficient walking must be learned through policy search, yet task-level objectives alone do not directly guide step-to-step coordination.
Instead, energy efficiency is commonly encouraged through effort- and smoothness-related penalties or explicit energy constraints \cite{FuEtAL2021,JeonEtAL2023,KimEtAL2024,HuangEtAL2026}.
These penalties and constraints target the overall energetic cost of a behavior without specifying how mechanical work should be organized across joints or across the transition between stance legs.
Although such terms remain useful, and the baseline objective shared by all conditions in this work includes an effort-related regularizer, they act at the level of the objective, where energy-related costs are balanced against competing objectives, such as velocity tracking and stability.
Consequently, overly strong regularization can degrade task performance \cite{JeonEtAL2023}.

A second and increasingly prominent line of work obtains this organization from human motion data rather than from hand-designed terms.
Imitation-based methods and adversarial motion-prior approaches use human motion-capture references to provide temporal and whole-body organization, which is difficult to specify exhaustively through reward design \cite{PengEtAL2018,PengEtAL2021,MahmoodEtAL2019}.
Such priors, however, score reference similarity in a kinematic feature space rather than actuator work.
Retargeting then matches joint configuration between bodies that differ in joint arrangement, segment inertia, and actuation.
In particular, human joint-power estimates include elastic tendon contributions that are not represented in the robot's actuator-work measure \cite{FarrisEtAL2012}.
As a result, similar kinematics do not imply similar actuator work or mechanical energy distribution.

One promising source of mechanics-informed guidance comes from passive dynamic walking (PDW).
Classical passive walkers can generate stable cyclic gaits on shallow downhill slopes through the interaction of gravity, inertia, impacts, and contact timing, without active actuation \cite{McGeer1990a,McGeer1990}, and subsequent limit-cycle walking studies characterized how such gaits reject disturbances without continuous control \cite{HobbelenEtAL2007}.
Powered passive-dynamic walkers further show that passive walking principles can inform efficient level-ground locomotion when actuation replaces the energetic role of downhill gravity \cite{CollinsEtAL2005}.
Although passive walkers differ substantially from modern high-degree-of-freedom humanoids, they suggest a useful training principle: mechanically efficient whole-body gait patterns may be more readily discovered when early policy search is conducted under conditions favorable to passive dynamics.
Inspired by this principle, we posit that temporarily modifying the training dynamics during early policy search to favor these conditions can bias learning toward robot-specific mechanically efficient coordination.

Based on this hypothesis, we propose passive-dynamic-walking-inspired, or PDW-inspired, dynamics guidance for humanoid RL (Fig.~\ref{fig:overview}).
During early training, a command-conditioned tilted-gravity field supplies slope-equivalent sagittal assistance while the collision geometry remains flat.
Curriculum-coupled objectives promote progression and discourage positive actuator work, and all PDW-specific guidance is withdrawn before final nominal-dynamics optimization.
Prior assistive-force curricula primarily use temporary support to facilitate skill acquisition \cite{ShiEtAL2023,CaoEtAL2025}.
Our study instead asks whether mechanically structured assistance can also bias the final unassisted policy toward a more economical gait, beyond any effect on acquisition speed.
The curriculum and an adversarial motion prior (AMP) therefore act on different aspects of learning: the former on the dynamics of gait search, the latter on reference-derived kinematic coordination \cite{PengEtAL2021}.

We evaluate the two separately and together in a $2\times2$ design on a Unitree G1 humanoid with 29 actuated degrees of freedom (DoF), through controlled sagittal and omnidirectional studies in simulation, followed by validation on the physical robot.

The main contributions of this work are threefold.
\begin{enumerate}
    \item \textbf{Framework.} We introduce PDW-inspired dynamics guidance that combines temporary slope-equivalent assistance with curriculum-coupled objectives and removes all PDW-specific guidance before final nominal-dynamics optimization.
    \item \textbf{Analysis.} Through controlled multi-seed studies, we show that the framework accelerates sustained stepping acquisition and separately yields a more economical converged gait, with the final mechanical-CoT reduction concentrated in lower positive actuator work.
    \item \textbf{Generalization.} We extend the framework to the complete omnidirectional command set and show in a $2\times2$ study that its benefit persists when a walking-specific motion prior is added, the combined condition attaining the lowest speed-matched cost of transport. A preliminary physical-robot evaluation reproduces the forward and lateral efficiency gains when the curriculum is combined with the motion prior.
\end{enumerate}

This article is organized as follows.
Section~\ref{sec:related} reviews related work, Section~\ref{sec:method} presents the proposed framework, and Section~\ref{sec:setup} the experimental setup.
Section~\ref{sec:results} reports the simulation and hardware results, Section~\ref{sec:discussion} discusses implications and limitations, and Section~\ref{sec:conclusion} concludes.
Complete implementation details and per-condition results appear in the appendices.

\section{Related Work}
\label{sec:related}
Prior work on legged locomotion has developed along two closely related lines: model-based approaches that use simplified dynamics and gait structure, and learning-based approaches that acquire feedback policies through trial-and-error interaction \cite{SuttonEtAL1998}.
Within the model-based tradition, PDW provides an important mechanical view of how efficient cyclic gait can emerge from favorable body dynamics \cite{McGeer1990a,CollinsEtAL2005}; within the learning-based tradition, economy has been addressed mainly through effort-aware regularization or constrained optimization \cite{FuEtAL2021,JeonEtAL2023,KimEtAL2024,HuangEtAL2026}, with motion priors supplying complementary guidance for coordinated motion \cite{PengEtAL2021,PengEtAL2018}.
Across both traditions, mechanical insight has been embodied in the machine, imposed by the deployed controller, or introduced into learning through the objective or a kinematic reference; the energetic loading under which a policy searches has received less attention.

\subsection{Passive Dynamic Walking and Efficient Gait Mechanics}
Classical humanoid locomotion control relies on simplified models, planned gait structures, and feedback laws designed from physical insight, using the zero-moment point and linear inverted pendulum model \cite{KajitaEtAL2003}, capture-point and foot-placement rules \cite{GriffinEtAL2017}, or hybrid zero dynamics \cite{WesterveltEtAL2007}.
These methods expose the physical structure of walking, but scaling them to high-degree-of-freedom humanoids with diverse commands and complex contacts remains difficult, which is why recent work has increasingly adopted learning-based controllers.

Within the same model-based tradition, passive dynamic walking shows that efficient bipedal gait can emerge from the natural dynamics of the body.
McGeer demonstrated that simple legged mechanisms can walk down a shallow slope using gravity, inertia, impacts, and contact timing, without active actuation \cite{McGeer1990a,McGeer1990}, and powered passive-dynamic walkers later achieved efficient level-ground walking when small amounts of actuation replaced the energetic role of downhill gravity \cite{CollinsEtAL2005}.
Virtual-slope approaches then showed that this energetic effect can be reproduced without an inclined surface: Dong et al. used controlled actuation to produce an effect analogous to downhill gravitational energy input in an underactuated biped \cite{DongEtAL2011}.
In contrast to using virtual-slope effects as part of the locomotion controller, our method introduces a command-conditioned tilted-gravity field only during policy training and removes it completely before the final nominal-dynamics training stage and deployment.

Passive dynamics is also closely related to human walking biomechanics.
Step-to-step transitions are a major contributor to the mechanical work and metabolic cost of walking, because the body center of mass must be redirected from one stance leg to the next \cite{DonelanEtAL,KuoEtAL2005}.
At the joint level, human walking draws its positive joint power mainly from the ankle and the hip, each contributing approximately 40--50\% of the total, while the knee contributes only 14--17\% and shows a largely negative profile; this composition is reported to be stable across walking speeds \cite{FarrisEtAL2012}.
However, classical passive walkers are low-dimensional mechanisms and do not solve the control problem faced by high-degree-of-freedom humanoids, which must track commands, reject disturbances, and operate on level ground.

\subsection{Reinforcement Learning for Legged Locomotion}
Reinforcement learning has become a widely used framework for legged locomotion because it can learn feedback controllers without prescribing full joint trajectories, contact schedules, or center-of-mass motions \cite{SuttonEtAL1998}.
Recent methods combine large-scale simulation, domain randomization, command-conditioned policies, terrain curricula, and sim-to-real transfer to train robust controllers for quadrupeds, bipeds, and humanoids \cite{RadosavovicEtAL2024,GuEtAL2024,LiEtAL2025a,RudinEtAL,SiekmannEtAL2021}.

This progress has also made reward and training design central to modern locomotion RL.
Rather than relying on a single task objective, most methods use composite objectives that combine command tracking with terms for stability, regularity, safety, energy use, and motion quality \cite{RadosavovicEtAL2024,GuEtAL2024,LiEtAL2025a,FuEtAL2021,JeonEtAL2023}.
Imitation-based and motion-prior methods can further guide policies toward natural whole-body motions by using reference motion data as an additional learning signal \cite{PengEtAL2018}.
Adversarial motion priors have been used in place of hand-designed reward terms and can yield energy-efficient gaits \cite{EscontrelaEtAL2022}, and subsequent work has addressed the training instability of the adversarial objective by changing the divergence measure or by combining multiple priors \cite{TangEtAL2024,VollenweiderEtAL2023}.
These techniques guide learning primarily through reward design or reference-motion distributions.
Beyond these approaches, curricula can also modify the physical assistance experienced during training.

Shi et al. introduced assistive-force curricula for learning reference-free bipedal motor skills, using temporary external forces to facilitate early exploration and removing them as the policy acquired the target behavior \cite{ShiEtAL2023}, and adaptive variants have since learned state-dependent support for high-dimensional humanoid skills \cite{CaoEtAL2025}.
Our method is related in its temporary use of external assistance but differs in structuring that assistance according to virtual-slope and passive-walking mechanics and in investigating whether it biases the final unassisted policy toward lower positive mechanical work.

\subsection{Energy-Efficient Locomotion Learning}
Energy efficiency in learned locomotion has commonly been addressed as part of reward or optimization design.
A common approach is to add effort-related penalties, such as torque, power, mechanical work, action change, or smoothness terms, to reduce unnecessary actuation while preserving tracking and balance \cite{FuEtAL2021,JeonEtAL2023}.
These penalties are simple to implement, but they introduce weights that must be tuned against other objectives, and poorly balanced reward terms can conflict with desired behaviors or make learning sensitive to reward scaling \cite{JeonEtAL2023}.

Recent work has therefore explored more structured formulations.
Potential-based reward shaping guides learning while reducing some sensitivity to reward design \cite{JeonEtAL2023}, smoothness-oriented approaches reduce jerky actions at deployment although they may require platform-specific tuning \cite{ChenEtAL2025b}, and ECO separates energy-related quantities from the task reward by formulating them as explicit inequality constraints enforced with a Lagrangian method \cite{HuangEtAL2026}.
These approaches improve the reward or optimization side of energy-efficient locomotion.

Energy efficiency has also been pursued through passive-dynamics-oriented design.
The Duke Humanoid combines a passive-dynamics-inspired 10-DoF hardware design with an RL controller that encourages the robot to exploit passive dynamics, lowering cost of transport in both simulation and hardware \cite{XiaEtAL2025}.
Efficient gait can therefore be promoted by shaping the mechanical and energetic conditions under which locomotion occurs, but such approaches retain passive-dynamics-oriented control mechanisms or specialized morphology during locomotion.

Across these lines, economy has been pursued by redesigning the robot, by imposing a passive walking template at run time, or by modifying reward and constraint terms; the present work instead changes the dynamics experienced during early policy learning.

\begin{figure*}[!t]
\centering
\includegraphics[width=\textwidth]{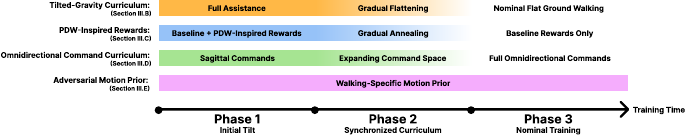}
\caption{Training schedule of the proposed framework. The three curriculum components and the adversarial motion prior are shown against training time, divided into the three phases used throughout this article. \emph{Phase 1}: the tilted-gravity field supplies full slope-equivalent assistance, the PDW-inspired reward terms are active alongside the baseline objective, and commands are sagittally dominant. \emph{Phase 2}: the tilt decays linearly and the PDW-inspired reward terms are scaled down by the same curriculum factor $\sigma(k)$, while the command distribution expands toward the full omnidirectional set. \emph{Phase 3}: all PDW-specific guidance has vanished, and the policy is optimized under nominal flat-ground dynamics, the baseline objective alone, and the complete command distribution. The adversarial motion prior, when enabled, is active throughout training and is not part of the curriculum. Fig.~\ref{fig:overview} illustrates the mechanical intuition for the tilted-gravity component.}
\label{fig:method_overview}
\end{figure*}

\section{Method}
\label{sec:method}
We consider flat-ground omnidirectional velocity tracking on an unmodified 29-DoF humanoid.
The deployment task, policy interface, and baseline locomotion objective are identical across all comparisons.
The proposed PDW-inspired curriculum modifies early training through three coupled components: (i)~command-conditioned virtual tilted gravity, (ii)~curriculum-coupled PDW-inspired rewards, and (iii)~a synchronized expansion from sagittally dominant to omnidirectional commands.
At the end of the curriculum, the tilted-gravity field and PDW-inspired reward terms vanish, while the command distribution reaches the complete omnidirectional task.

We additionally use a walking-specific adversarial motion prior as a motion-quality objective that operates orthogonally to the curriculum.
Treating it as a separate experimental factor lets us both establish the curriculum's contribution without AMP and quantify how the two combine; when enabled, AMP remains active throughout training.
Fig.~\ref{fig:method_overview} summarizes how these components are scheduled over training.

\subsection{Task and Baseline Objective}
At control step $t$, the planar velocity command is
\begin{equation}
\mathbf{c}_t = \left( \mathbf{c}_{v,t}, \omega_{z,t}^{*} \right), \qquad
\mathbf{c}_{v,t} = \left( v_{x,t}^{*}, v_{y,t}^{*} \right),
\end{equation}
where $\mathbf{c}_{v,t}$ is the translational velocity command in the yaw-aligned body frame and $\omega_{z,t}^{*}$ is the commanded yaw rate.
The commanded translational speed and direction are
\begin{equation}
s_t^{*} = \left\| \mathbf{c}_{v,t} \right\|_2, \qquad
\hat{\mathbf{c}}_{v,t} = \frac{\mathbf{c}_{v,t}}{\max\left( s_t^{*}, \epsilon_v \right)},
\end{equation}
where $\epsilon_v>0$ prevents division by zero.
The measured planar base velocity in the same yaw-aligned frame is denoted by $\mathbf{v}_{xy,t}^{b}$.

Let $r_t^{\mathrm{base}}$ denote the baseline locomotion reward.
Following common practice in learning-based legged and humanoid locomotion, it combines linear- and angular-velocity tracking with regularization for body stability, posture, joint motion, action smoothness, undesirable contacts, and actuator effort \cite{RudinEtAL,FuEtAL2021,RadosavovicEtAL2024,JeonEtAL2023}.
The same baseline objective is used throughout training and across all comparisons.
Its complete list of terms and weights is given in Table~\ref{tab:impl_basereward}.
The remainder of this section defines only the components introduced by the proposed framework.

\subsection{Command-Conditioned Tilted-Gravity Curriculum}
\label{subsec:tilt}
\subsubsection{Curriculum Schedule}
Let $k$ denote the curriculum counter, measured in episode-equivalent units; its mapping to PPO policy updates is listed in Table~\ref{tab:impl_pdw}.
The transition progress and remaining assistance are defined as
\begin{equation}
\beta(k) = \operatorname{clip}\left( \frac{k-K_w}{K_t-K_w}, 0, 1 \right), \qquad
\sigma(k) = 1-\beta(k),
\end{equation}
where $K_w$ marks the end of the full-assistance stage and $K_t=200$ curriculum episodes marks the end of the transition.
The omnidirectional study uses $K_w=150$ curriculum episodes; the value used for the forward mechanism study is listed separately in Table~\ref{tab:impl_pdw}.
The nominal tilt angle is
\begin{equation}
\theta(k) = \theta_{\mathrm{init}}\, \sigma(k),
\end{equation}
where $\theta_{\mathrm{init}}$ is the forward initial-tilt magnitude.
Thus, the full tilt is maintained for $k<K_w$, decreases linearly over $K_w\leq k<K_t$, and becomes zero for $k\geq K_t$.
The initial tilt is treated as a design parameter rather than a physical constant; its default value is $\theta_{\mathrm{init}}=5^{\circ}$ (Table~\ref{tab:impl_pdw}) and its sensitivity is evaluated in the initial-angle ablation (Table~\ref{tab:tilt_sweep}).

\subsubsection{Direction-Dependent Tilted Gravity}
A downhill slope supplies gravitational energy along its direction of descent.
In the proposed curriculum, this assistance is used to facilitate fore--aft step-to-step progression.
The virtual tilt is therefore aligned with the robot's yaw-aligned sagittal axis.
Lateral locomotion is introduced through the command curriculum in Section~\ref{subsec:omnicmd} rather than by rotating the downhill direction toward each lateral command.

The sagittal alignment of the translational command is
\begin{equation}
u_t = \frac{v_{x,t}^{*}}{\max\left( s_t^{*}, \epsilon_v \right)}.
\end{equation}
We define the direction-dependent coefficient
\begin{equation}
a(u_t) = \left[ u_t \right]_{+} - \rho \left[ -u_t \right]_{+}, \qquad
\rho = \frac{\sin\theta_{\mathrm{back}}}{\sin\theta_{\mathrm{init}}},
\end{equation}
where $[x]_{+} = \max(x,0)$ and the backward-to-forward tilt ratio $\rho$ is set by a second design parameter $\theta_{\mathrm{back}}$, the maximum backward tilt magnitude.
The effective tilt angle is then
\begin{equation}
\theta_{\mathrm{eff},t} = \arcsin\left( \sin\theta(k)\, a(u_t) \right).
\end{equation}
At full assistance a pure forward command therefore receives the tilt $\theta_{\mathrm{init}}=5^{\circ}$, a pure backward command an oppositely directed tilt of magnitude $\theta_{\mathrm{back}}=3^{\circ}$, a diagonal command an assistance proportional to its sagittal component, and a purely lateral or stationary command none.

Let $\mathbf{e}_{x,t}^{W}$ be the robot's yaw-aligned forward direction expressed in the world frame, let $\mathbf{e}_{z}^{W}$ be the world upward direction, and let $\mathbf{g}_0 = -g\,\mathbf{e}_{z}^{W}$ be the nominal gravity vector, where $g$ is the gravitational acceleration.
The tilted gravity vector is
\begin{equation}
\tilde{\mathbf{g}}_t = g \sin\theta_{\mathrm{eff},t}\, \mathbf{e}_{x,t}^{W} - g \cos\theta_{\mathrm{eff},t}\, \mathbf{e}_{z}^{W},
\end{equation}
with $\left\|\tilde{\mathbf{g}}_t\right\|_2=g$.
For each rigid link $i$ with mass $m_i$, we apply
\begin{equation}
\mathbf{F}_{i,t}^{\mathrm{ext}} = m_i \left( \tilde{\mathbf{g}}_t - \mathbf{g}_0 \right), \qquad
\boldsymbol{\tau}_{i,t}^{\mathrm{ext}} = \mathbf{0}.
\end{equation}
The projected-gravity observation is computed using $\tilde{\mathbf{g}}_t$, so that the policy observes a gravity direction consistent with the applied force.

On a locally planar slope, gravity has a tangential component $g\sin\theta$ and a normal component $g\cos\theta$ relative to the support plane; the tilted-gravity field reproduces both for the prescribed tilt while keeping the collision geometry flat.
For a fixed heading the two are physically equivalent---rotating the world until the tilted gravity points straight down recovers a true slope of the same angle---so the field reproduces the local mechanics of slope walking without inclined geometry.
The command-conditioned alignment with the commanded direction generalizes a single fixed slope, and the flat geometry isolates gravity-assisted gait discovery from terrain effects such as height variation, slope transitions, and surface irregularities.

Keeping the geometry flat also supports massively parallel training, since each environment can use an independent tilt magnitude and direction without requiring a separate terrain mesh or a simulator-wide gravity setting.

\subsection{Curriculum-Coupled PDW-Inspired Rewards}
Tilted gravity changes the energetic conditions of early learning, but it does not determine how the policy should distribute actuator work, maintain progression, or organize contacts.
We therefore augment the baseline objective as
\begin{equation}
r_t^{\mathrm{task}} = r_t^{\mathrm{base}} + r_t^{\mathrm{PW}} + r_t^{\mathrm{VB}} + r_t^{\mathrm{air}}.
\end{equation}
Each additional term is multiplied by $\sigma(k)$ and vanishes when the tilted-gravity curriculum ends.
The baseline locomotion reward remains active throughout training.

\subsubsection{Positive-Work Penalty}
Let $P_{j,t} = \tau_{j,t}\, \dot{q}_{j,t}$ denote the instantaneous mechanical power of joint $j$.
We penalize only its positive component, so that the cumulative reward discourages positive mechanical work:
\begin{equation}
r_t^{\mathrm{PW}} = -w_{\mathrm{PW}}\, \sigma(k) \sum_{j=1}^{29} \left[ P_{j,t} \right]_{+}.
\end{equation}
Positive and negative joint power serve different mechanical roles: positive work injects energy while negative work absorbs it, and both occur during step-to-step transitions \cite{DonelanEtAL2002}.
Negative power is likewise needed to stabilize a humanoid torso, and passive-dynamic walkers rely on comparable absorption at knee strike \cite{McGeer1990} and at the hip once an upper body is added \cite{WisseEtAL2007}.
An absolute-power penalty $\sum_j |P_{j,t}|$ would penalize this stabilizing absorption equally with active energy generation; $r_t^{\mathrm{PW}}$ instead discourages only the replacement of tilted-gravity assistance with unnecessary positive actuator work, leaving energy absorption and any effort regularizer in $r_t^{\mathrm{base}}$ unchanged.
The positive mechanical work reported as an evaluation metric is defined separately in Section~\ref{subsec:metrics}.

\subsubsection{Velocity-Band Reward}
Classical passive-dynamic walkers can converge to a stable periodic gait when their state lies within the basin of attraction of a walking limit cycle \cite{McGeer1990a,GarciaEtAL1998}.
Such a cycle emerges from the interaction of gravity, inertia, impacts, and contact timing rather than from an explicitly prescribed gait phase.
Motivated by this property, we provide a broad progression objective during the tilted-gravity stage:
\begin{equation}
r_t^{\mathrm{VB}} = w_{\mathrm{VB}}\, \sigma(k)\, \mathbf{1}\!\left[ v_{\min} \leq \mathbf{v}_{xy,t}^{b} \cdot \hat{\mathbf{c}}_{v,t} \leq \eta_{\mathrm{VB}}\, s_t^{*} \right],
\end{equation}
where $v_{\min}$ is the band lower threshold and $\eta_{\mathrm{VB}}$ is the command-relative upper bound.
The upper bound is command-relative because the omnidirectional command set spans a wide range of speeds for which a fixed cap would be ill-defined; the purely sagittal forward study, in which the projection reduces to $v_{x,t}^{b}$, uses a fixed absolute cap instead.
Stationary commands are excluded whenever the term is active.
The lower bound withholds reward for stationary or reversed motion and the upper bound prevents extra reward from accelerating excessively under tilted gravity.
Because the reward is constant within the admissible band, it encourages command-aligned progression without prescribing an exact speed, gait phase, cadence, or step length.

The band does not guarantee convergence to a limit cycle; it only provides a permissive region in which periodic coordination can emerge.
As $\sigma(k)$ decreases, the band reward is removed and the baseline command-tracking objective becomes dominant, shifting optimization toward accurate velocity tracking under nominal flat-ground dynamics.

\subsubsection{Bilateral-Flight Penalty}
During early training, the policy may produce repeated simultaneous flight of both feet instead of alternating walking contacts.
Let $\chi_{L,t}$ and $\chi_{R,t}$ denote the left- and right-foot contact indicators.
We define
\begin{equation}
r_t^{\mathrm{air}} = -w_{\mathrm{air}}\, \sigma(k)\, \mathbf{1}\!\left[ \chi_{L,t}=0 \land \chi_{R,t}=0 \right] \mathbf{1}\!\left[ s_t^{*} < v_{\mathrm{thr}} \right].
\end{equation}
This term penalizes bilateral flight under low-speed commands, for which a flight phase is not required by the task.
This degeneracy was observed specifically under backward commands during assisted low-speed training, whereas forward and lateral commands did not induce it; the term therefore targets a direction-specific failure mode and is not needed for purely forward locomotion.
The coefficients and thresholds of all PDW-inspired reward terms are listed in Table~\ref{tab:impl_pdw}.

\subsection{Omnidirectional Command Curriculum}
\label{subsec:omnicmd}
\begin{figure}[!t]
\centering
\includegraphics[width=\columnwidth]{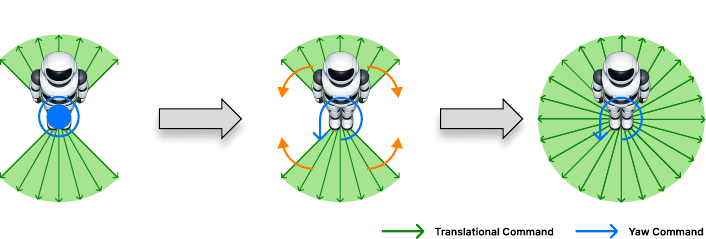}
\caption{Synchronized expansion of the translational command sectors and the yaw range over the curriculum. \emph{Left} ($k<K_w$): commands are restricted to two sagittal sectors spanning $\pm 45^{\circ}$ about the forward and backward directions, and no yaw command is issued. \emph{Middle} ($K_w \leq k < K_t$): each sector expands continuously toward $\pm 90^{\circ}$ while the yaw range opens. \emph{Right} ($k \geq K_t$): the union of the two sectors covers the full $360^{\circ}$ translational command range at the final yaw range. Green arrows denote admissible translational command directions; blue denotes the commanded yaw rate.}
\label{fig:command_curriculum}
\end{figure}
Fig.~\ref{fig:command_curriculum} illustrates the synchronized expansion from sagittal to omnidirectional locomotion.
The curriculum reflects the biomechanical asymmetry between forward progression and lateral balance: sagittal walking can exploit pendular exchange and passive limb dynamics, whereas mediolateral balance requires active regulation---the dominant unstable mode of a passive walking model lies in lateral motion, and humans regulate it actively through mechanisms including lateral foot placement \cite{BaubyEtAL2000}.
Sustained sideways walking is also slower and more energetically costly than forward walking, as it requires repeated lateral acceleration and deceleration \cite{HandfordEtAL2014}.

Accordingly, the method does not treat lateral locomotion as a $90^{\circ}$-rotated downhill task: the tilted gravity remains sagittal and provides no direct assistance to a pure lateral command.
Instead, the curriculum treats sustained lateral translation as an extension of the mediolateral balance action already required during forward walking.
Early training establishes a repeatable sagittal gait, after which the command range expands so that the learned lateral balance response develops into sustained lateral motion.

The translational command heading $\psi_t^{*} = \operatorname{atan2}( v_{y,t}^{*}, v_{x,t}^{*} )$ is drawn from two sectors centered on the forward and backward directions, each of half-width
\begin{equation}
\psi_{\max}(k) = \frac{\pi}{4} + \frac{\pi}{4} \beta(k).
\end{equation}
The admissible heading support therefore grows from $180^{\circ}$ at $k<K_w$ to the full $360^{\circ}$ at $k\geq K_t$ (Fig.~\ref{fig:command_curriculum}).

Yaw commands are introduced over the same transition, with $\omega_{z,t}^{*}$ drawn from $\left[ -\omega_{\max}(k), \omega_{\max}(k) \right]$ and
\begin{equation}
\omega_{\max}(k) = \beta(k)\, \omega_{\max}^{\mathrm{final}}.
\end{equation}
Thus, the full-assistance stage uses sagittal command sectors and zero yaw command.
During the transition, the virtual tilt is removed while the translational sectors and yaw range expand.
Training continues to $k=K_{\mathrm{final}}\approx360$ curriculum episodes, leaving approximately 44\% of training under fully nominal dynamics after the tilt is withdrawn at $K_t$.

\subsection{Walking-Specific Adversarial Motion Prior}
\label{subsec:amp}
Natural humanoid walking requires temporal coordination among the torso, limbs, and contacts that is difficult to specify exhaustively with manually designed reward terms.
Human motion-capture data provide a data-driven prior over such coordinated transitions.
We therefore incorporate Adversarial Motion Priors (AMP) as a walking-specific motion-quality objective \cite{PengEtAL2021}.

The reference distribution contains 127 retargeted walking clips selected from AMASS \cite{MahmoodEtAL2019}.
Restricting the reference set to walking motions makes the learned prior specific to locomotion rather than to general whole-body motion.
Let $\Phi(s_t)$ denote the yaw- and translation-invariant motion feature used by the discriminator, and let $D_{\xi}$ be a discriminator that distinguishes reference from policy transitions $\left( \Phi(s_t), \Phi(s_{t+1}) \right)$.
Following AMP, the motion-prior reward is
\begin{equation}
r_t^{\mathrm{AMP}} = \max\left( 0,\, 1 - \tfrac{1}{4}\left( D_{\xi}\left( \Phi(s_t), \Phi(s_{t+1}) \right) - 1 \right)^2 \right).
\end{equation}
The complete training reward is
\begin{equation}
r_t = 0.9\, r_t^{\mathrm{task}} + 0.1\, r_t^{\mathrm{AMP}}.
\end{equation}
The $0.9{:}0.1$ weighting keeps command-conditioned locomotion as the primary objective and uses AMP as auxiliary guidance for temporal and kinematic coordination.
The motion feature, discriminator architecture, and optimization settings are reported in Table~\ref{tab:impl_amp}.

Unlike the PDW-inspired terms, $r_t^{\mathrm{AMP}}$ is not coupled to $\sigma(k)$: it remains active after the tilted-gravity field and the curriculum-coupled objectives have been removed.
We therefore treat AMP as a motion-quality factor separate from the PDW mechanism and evaluate the two independently and in combination (Section~\ref{subsec:omni}).

\section{Experimental Setup}
\label{sec:setup}

\subsection{Robot and Simulation}
The robot is a 29-DoF Unitree G1.
The nominal robot model has a mass of $m = 33.34$~kg, which is used as the mass in all reported cost-of-transport values. Because this mass cancels in every relative comparison, the reported percentage reductions are invariant to its value.
Training runs in Isaac Lab on Isaac Sim (PhysX) at a $50$~Hz control rate ($\Delta t = 0.02$~s), with 1024 parallel environments and 20~s episodes.
Actions are 29 joint position targets tracked by implicit PD actuators.
The full simulation, control, and actuator-gain settings are listed in Tables~S1 and~S2.

The terrain asset contains sloped cells, but the terrain-level curriculum is disabled and the initial level is fixed at zero, so every environment remains on a flat cell for the entire run.

\subsection{Policy and Optimization}
\label{subsec:policy}
The policy is recurrent: a 5-step stack of base angular velocity, projected gravity, velocity command, joint positions and velocities, and the previous action enters a single-layer LSTM (256 units) followed by an MLP with hidden sizes $[256, 128]$; the critic additionally receives base linear velocity.
Baseline and the proposed method use identical architectures.
Appendix~\ref{app:arch} shows that replacing the recurrent core with a feed-forward policy preserves the efficiency result, so the reported gains do not depend on recurrence.

We train with PPO for 15{,}000 updates (368.6M environment steps).
The network dimensions and the full PPO hyperparameters are listed in Table~\ref{tab:impl_ppo}.

\subsection{Training Configuration}
Friction, torso mass, periodic base-velocity pushes, and reset pose and velocity are randomized identically in every condition; Table~\ref{tab:impl_sim} lists the ranges.
At deployment the command ranges are $v_x^{*} \in [-0.8, 2.0]$~m/s, $v_y^{*} \in [-0.8, 0.8]$~m/s, $\omega_z^{*} \in [-0.8, 0.8]$~rad/s, with dead zones of 0.1~m/s and 0.05~rad/s, resampled every 10~s and smoothed with an exponential filter.
The adversarial motion prior is defined in Section~\ref{subsec:amp}; its reference set, motion features, discriminator architecture, and optimization settings are listed in Table~\ref{tab:impl_amp}.

Unless stated otherwise, all conditions share the deployment task, policy interface, observation space, action space, and baseline reward (specified term-by-term in Table~\ref{tab:impl_basereward}); they differ only in the training-time components under study.

\subsection{Evaluation Metrics}
\label{subsec:metrics}
All metrics are computed after the curriculum has ended, under nominal flat-ground dynamics with no tilted gravity and no curriculum-coupled reward terms active.
Let $\tau_{j,t}$ and $\dot q_{j,t}$ denote the torque and velocity of joint $j$ at step $t$, $g = 9.81$~m/s\textsuperscript{2}, $d$ the distance travelled in an episode, and $m = 33.34$~kg the robot mass from Section~\ref{sec:setup}.

\subsubsection{Cost of Transport}
The primary economy metric, dimensionless:
\begin{equation}
\mathrm{CoT} = \frac{\sum_t \sum_{j=1}^{29} \left| \tau_{j,t} \dot q_{j,t} \right| \Delta t}{m\, g\, d}.
\end{equation}
Episodes with traveled distance below $0.5$~m are excluded, since the ratio diverges as $d$ approaches zero; episodes in which the robot fails to make progress are instead counted in the stuck rate defined below.

\subsubsection{Positive and Negative Mechanical Work}
Per-joint clipping before summation:
\begin{equation}
W^{+} = \sum_t \sum_j \left[ \tau_{j,t} \dot q_{j,t} \right]_{+} \Delta t,
\end{equation}
\begin{equation}
W^{-} = \sum_t \sum_j \left| \min \left( \tau_{j,t} \dot q_{j,t}, 0 \right) \right| \Delta t .
\end{equation}
By construction $W^{+} + W^{-} = \sum_t \sum_j |\tau_{j,t} \dot q_{j,t}| \Delta t$, so CoT decomposes exactly into a positive and a negative component, $\mathrm{CoT} = \mathrm{CoT}^{+} + \mathrm{CoT}^{-}$, with $\mathrm{CoT}^{\pm} = W^{\pm} / (m g d)$.
This decomposition is what allows the mechanism analysis in Section~\ref{subsec:work}: an improvement in CoT can be attributed to its positive and negative parts additively.
We also report the \emph{positive work ratio} $W^{+} / (W^{+} + W^{-})$, and \emph{positive work per meter} $W^{+}/d$ [J/m], which is independent of the mass normalization.

\subsubsection{Velocity-Tracking Error}
For the sagittal study, the time-averaged absolute error of the body-frame forward velocity against its command, $\frac{1}{T}\sum_t | v_{x,t}^{b} - v_{x}^{*} |$.
For the omnidirectional study, the corresponding planar norm $\frac{1}{T}\sum_t \| \mathbf{v}_{xy,t}^{b} - \mathbf{c}_{v} \|_2$.

\subsubsection{Gait Symmetry}
The ratio of mean stance durations, $\min(\bar T_{L}, \bar T_{R}) / \max(\bar T_{L}, \bar T_{R})$, where stance intervals are detected from vertical ankle contact force.
A value of 1 indicates left and right stance phases of equal duration.

\subsubsection{Stuck and Tracking Rates}
The fraction of episodes in which the robot fails to make progress, and the complementary fraction in which it follows the commanded direction.
These distinguish \emph{inability to perform the command} from \emph{inefficiency while performing it}, which Section~\ref{subsec:omni} shows is essential for interpreting the omnidirectional results.

\subsubsection{Speed-Matched Comparison}
CoT depends on walking speed, and policies do not realize commanded speed identically: a policy that systematically undershoots its command is evaluated at a different point on the CoT--speed curve than one that tracks it.
Comparing condition means over a command distribution therefore conflates economy with speed realization.

Wherever conditions differ in realized speed, we report \emph{speed-matched} values in addition to raw means.
Each episode is assigned to a bin by its \emph{realized} mean speed $d / (T \Delta t)$; bins are taken over the speed range common to all conditions being compared; and each condition's metric is averaged within bins and then across bins with equal weight.
This removes differences in the speed \emph{mixture} between conditions.
Confidence statements over episodes use a bootstrap; where multiple seeds exist, we report variability across seeds, which is the quantity relevant to claims about the method rather than about one trained policy.

\subsubsection{Joint-Level Work and Phase-Resolved Joint Power}
\label{subsubsec:jointmetrics}
For the omnidirectional $2\times2$ comparison we additionally characterize how mechanical work is organized among the sagittal leg joints during forward locomotion, using two analyses whose means and standard deviations are taken across the three independently trained policy seeds.

The first resolves the \emph{composition} of positive actuator work.
Per-joint positive work $W_{j}^{+} = \sum_t [\tau_{j,t}\dot q_{j,t}]_{+}\Delta t$ is summed over the left and right members of each of the ankle-pitch, knee, and hip-pitch groups, and each group's share is normalized by the sum over these three sagittal groups only rather than over all 29 actuators, so the analysis reports relative composition and not absolute work.
Shares are computed per episode under forward commands, averaged within $0.1$~m/s realized-speed bins over the common range $0.5$--$1.4$~m/s, and then averaged across bins with equal weight.

The second resolves \emph{when} in the gait cycle that work is done, which integrated shares cannot show.
It uses a separate evaluation with forward-only commands, $v_x \sim \mathcal{U}(0.4, 1.6)$~m/s and $v_y$ and yaw rate held at zero, whereas the work-share analysis draws its forward commands from the full omnidirectional distribution.
Strides are segmented from one touchdown of a foot to its next with left and right limbs pooled, resampled to a common phase grid, and binned by realized speed over $0.5$--$1.4$~m/s; segmentation criteria, rejection thresholds, and binning are given in Appendix~\ref{app:joint}.

Ankle push-off work is the positive ankle work in the final $30\%$ of stance and knee weight-acceptance work is the knee work in the first $15\%$ of stance; both windows are fractions of stance and therefore scale with each condition's own toe-off.

\subsection{Physical-Robot Deployment and Hardware Distance Recovery}
\label{subsec:hardware}
The trained policies are deployed on a physical 29-DoF Unitree G1 using the same 50~Hz PD position-control interface as in simulation, at the nominal deployed mass $m=33.34$~kg.
Four policies are evaluated in a $2\times2$ design over $\{\mathrm{Baseline},\mathrm{Ours}\}\times\{\text{no AMP},\text{AMP}\}$, using a single deployed policy per condition.
Three command directions are tested---forward ($1.2$~m/s), backward ($0.8$~m/s), and lateral ($0.8$~m/s, left and right pooled)---with several steady-state traversals per cell ($n=2$--$6$); for each traversal the initial velocity rise and a short end margin are discarded and metrics are computed over the steady-state segment only.
A traversal is a constant-command segment of $4.8$--$8.8$~m, $6.8$~m and $10.1$~s on average, so the reported hardware cost of transport aggregates 38 traversals and 258~m of measured steady-state walking.

On hardware the base position is not directly observable, so the travelled distance and speed used for cost of transport are obtained by replaying the joint angles and IMU signals logged on the real robot through the simulator's forward kinematics. Hardware $\mathrm{CoT}$ and its positive-work component $\mathrm{CoT}^{+}$ are then computed from these quantities using the definitions of Section~\ref{subsec:metrics}. Applying the same replay to simulation logs, where the true base displacement is known, recovers 91--100\% of the travelled distance across the evaluated conditions, so the procedure slightly underestimates distance and therefore biases the reported $\mathrm{CoT}$ upward.

\section{Experimental Results}
\label{sec:results}
The experiments address four questions.
First, does the proposed training scheme reduce mechanical cost of transport during forward locomotion without degrading command tracking?
Second, is the reduction explained by lower positive actuator work, as intended by the reward design?
Third, do the VB-guided training regime and tilted-gravity assistance affect when sustained stepping first emerges during training, and is their effect on converged economy separable from that timing?
Fourth, does the coordination developed under sagittal assistance extend to unassisted omnidirectional locomotion, how does it interact with AMP, and what joint-level organization accompanies the motion-prior effect?

The forward-locomotion study provides controlled component ablations and a mechanical-work analysis that support the main energy-efficiency and work-attribution claims, whereas the omnidirectional study evaluates whether the complete command-conditioned framework extends to the deployment task and how it interacts with AMP.
All evaluations are conducted after the tilted-gravity field and curriculum-coupled reward terms have been removed.

\subsection{Forward-Locomotion Mechanism Study}
\label{subsec:forward}
We first examine how tilted-gravity training, the velocity-band objective, and the positive-work penalty affect forward locomotion, where the mechanical role of the assistance is most directly expressed.
The forward study uses the same component structure as the omnidirectional framework, but with commands restricted to the sagittal task and with the bilateral-flight penalty inactive, as that term targets a backward-command failure mode that does not arise under forward locomotion; the full-assistance duration used for each study is listed separately in Table~\ref{tab:impl_pdw}.

Five conditions are compared, as summarized in Table~\ref{tab:conditions}.

\begin{table}[!t]
\caption{Conditions of the Forward-Locomotion Mechanism Study}
\label{tab:conditions}
\centering
\begin{tabular}{lccc}
\hline
Condition & Tilted gravity & Velocity band & Positive-work \\
 &  & (VB) & penalty (PW) \\
\hline
Flat (baseline)          & ---          & ---          & ---          \\
Flat + VB                & ---          & $\checkmark$ & ---          \\
Flat + VB + PW           & ---          & $\checkmark$ & $\checkmark$ \\
Slope + VB               & $\checkmark$ & $\checkmark$ & ---          \\
\textbf{Slope + VB + PW} & $\checkmark$ & $\checkmark$ & $\checkmark$ \\
\hline
\end{tabular}
\end{table}

\begin{figure}[!t]
\centering
\includegraphics[width=\columnwidth]{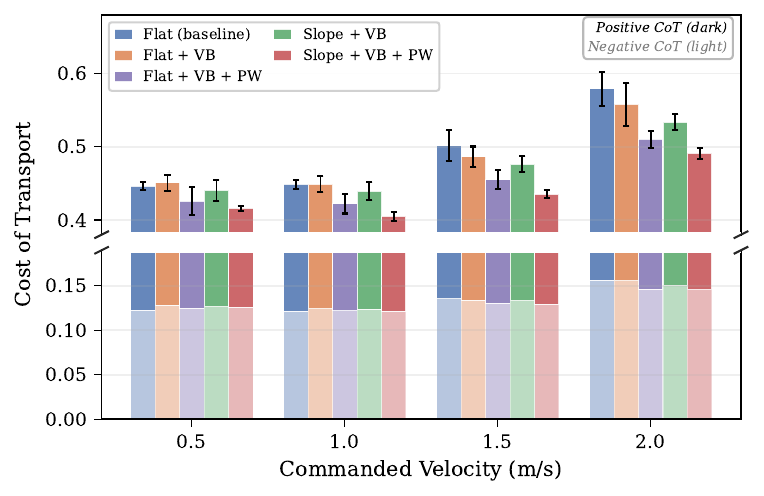}
\caption{Cost of transport on the sagittal task across commanded speeds (mean over 5 seeds; error bars show the S.D.\ of total CoT). Total bar height gives the total CoT, split into a positive component (dark) and a negative component (light). The full curriculum (Slope + VB + PW) attains the lowest CoT at every speed and lowers it primarily by shrinking the positive component, while the negative component stays essentially unchanged across all conditions and speeds.}
\label{fig:cot_decomp}
\end{figure}

Each condition is trained with five seeds $\{42,123,7,0,256\}$ and evaluated at commanded speeds of 0.5, 1.0, 1.5, and 2.0~m/s.
Fig.~\ref{fig:cot_decomp} reports mean CoT across independently trained seeds, with each bar decomposed into positive and negative components and error bars showing the standard deviation of total CoT.
Statistical comparisons use paired $t$-tests against the baseline over seeds.

The full condition reduces CoT by 6.8\%, 9.7\%, 13.2\%, and 15.2\% at 0.5, 1.0, 1.5, and 2.0~m/s, respectively, and the reduction is statistically significant at every speed (paired $t$-test vs.\ baseline, $p<0.05$; full per-condition values in Table~\ref{tab:appd1}).
The improvement increases over the evaluated speed range, consistent with the increasing step-to-step redirection demands of faster walking.

The velocity band alone does not significantly change CoT relative to the baseline.
Adding the positive-work penalty to Flat + VB produces a substantial reduction at intermediate and high speeds, while Slope + VB yields a smaller reduction.
Their combination gives the lowest CoT at all evaluated speeds.
At 1.0~m/s, for example, the full condition reduces CoT by 9.7\%, compared with 5.9\% for Flat + VB + PW and 2.1\% for Slope + VB.

The CoT reduction is not accompanied by a measurable loss in command tracking.
At 1.0~m/s, velocity-tracking error is $0.0850\pm0.0053$~m/s for the baseline and $0.0881\pm0.0097$~m/s for the full condition, and no significant difference is observed across the evaluated speeds.
Gait symmetry remains within 0.93--0.98 for all principal conditions.
Push-recovery survival is saturated near unity and is therefore not used to support comparative claims.

Training Slope + PW without the velocity band does not reproduce the full method: the positive-work penalty alone can favor degenerate low-work behaviors---in our runs, either a one-leg progression with large tracking error or walking at substantially higher CoT.
The velocity band prevents this by rewarding command-aligned progression without prescribing an exact speed or contact schedule.

\subsection{Mechanical-Work Analysis}
\label{subsec:work}
The positive-work penalty is intended to prevent the policy from replacing gravitational assistance with active energy injection while leaving stabilizing energy absorption unpenalized.
If this mechanism is realized, the CoT reduction should arise primarily from $\mathrm{CoT}^{+}$, with comparatively little change in $\mathrm{CoT}^{-}$.

\begin{table}[!t]
\footnotesize
\caption{Mechanical-Work Decomposition, $\mathrm{CoT}^{\pm}$ \\ (Mean $\pm$ S.D., 5 Seeds)}
\label{tab:appd2}
\centering
\setlength{\tabcolsep}{3.5pt}
\begin{tabular}{lcccc}
\hline
Condition & $1.0\,\mathrm{CoT}^{+}$ & $1.0\,\mathrm{CoT}^{-}$ & $2.0\,\mathrm{CoT}^{+}$ & $2.0\,\mathrm{CoT}^{-}$ \\
\hline
Flat (baseline) & $0.3274$ & $0.1213$ & $0.4229$ & $0.1562$ \\
 & ${\pm}0.0065$ & ${\pm}0.0053$ & ${\pm}0.0205$ & ${\pm}0.0095$ \\
Flat + VB & $0.3250$ & $0.1241$ & $0.4022$ & $0.1557$ \\
 & ${\pm}0.0083$ & ${\pm}0.0031$ & ${\pm}0.0202$ & ${\pm}0.0099$ \\
Flat + VB + PW & $0.3003$ & $0.1219$ & $0.3642$ & $0.1461$ \\
 & ${\pm}0.0152$ & ${\pm}0.0029$ & ${\pm}0.0143$ & ${\pm}0.0037$ \\
Slope + VB & $0.3158$ & $0.1234$ & $0.3840$ & $0.1499$ \\
 & ${\pm}0.0074$ & ${\pm}0.0053$ & ${\pm}0.0090$ & ${\pm}0.0020$ \\
\textbf{Slope + VB + PW} & $\mathbf{0.2837}$ & $0.1214$ & $\mathbf{0.3454}$ & $0.1456$ \\
 & ${\pm}0.0029$ & ${\pm}0.0068$ & ${\pm}0.0076$ & ${\pm}0.0066$ \\
\hline
\multicolumn{5}{l}{\footnotesize Commanded speeds in m/s. VB: velocity band; PW: positive-work penalty.} \\
\end{tabular}
\end{table}

The full condition lowers $\mathrm{CoT}^{+}$ by 13.3\% at 1.0~m/s and 18.3\% at 2.0~m/s (Fig.~\ref{fig:cot_decomp}; per-speed values in Table~\ref{tab:appd2}).
At 1.0~m/s the change in $\mathrm{CoT}^{-}$ is negligible, so essentially all of the total reduction is attributable to $\mathrm{CoT}^{+}$; at 2.0~m/s, $\mathrm{CoT}^{+}$ accounts for approximately 88\% of the total reduction.
This pattern shows that the policy does not obtain the improvement simply by shifting work into the negative-power branch.
The unchanged tracking error in Section~\ref{subsec:forward} rules out reduced task execution as the primary explanation.
Positive work per meter, $W^{+}/d$, is reported in Table~\ref{tab:tilt_sweep} as an interpretable dimensional quantity rather than as an independent cross-check, since it differs from $\mathrm{CoT}^{+}$ only by the constant factor $mg$.
These results are consistent with the intended positive-work mechanism: the curriculum reduces the positive mechanical work required to realize the same forward command while leaving energy absorption essentially untouched.

\subsection{Curriculum Sensitivity and Control Experiments}
\label{subsec:sensitivity}

\subsubsection{Emergence of Sustained Stepping}
We next examine whether the CoT reduction can be explained simply by faster acquisition of stepping.
Tilted-gravity assistance may lower the difficulty of early exploration, causing stepping to emerge earlier without necessarily changing the mechanics of the converged gait.
We therefore compare the training update at which sustained stepping first appears while holding the reward terms and their schedules fixed.

A step event is an air-to-contact transition of either foot.
The primary emergence criterion is the first update at which the fraction of episodes containing at least five step events exceeds 0.99, with mean step-event counts per episode as supporting measures; thresholds and smoothing are given in Appendix~\ref{app:forward}.
Every condition includes five independently trained seeds, and comparisons are paired over seeds.

\begin{figure}[!t]
\centering
\includegraphics[width=\columnwidth]{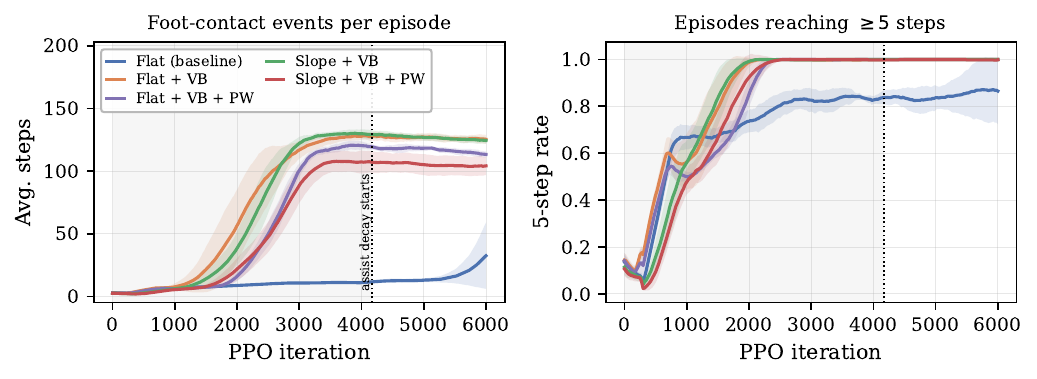}
\caption{Emergence of sustained stepping during training (mean $\pm$ s.d.\ over five seeds). Left: mean foot-contact events per episode. Right: fraction of episodes with at least five step events. The horizontal axis is the PPO policy-update index. The flat and tilted-gravity variants with matched reward terms follow similar trajectories, whereas the conventional baseline reaches sustained stepping substantially later.}
\label{fig:gait_emergence}
\end{figure}

The isolating comparisons for tilted gravity are Slope + VB versus Flat + VB and Slope + VB + PW versus Flat + VB + PW.
Within each pair, the reward terms and their time schedules are identical, and only the tilted-gravity field differs.
Fig.~\ref{fig:gait_emergence} shows the training curves for these comparisons.

The conventional baseline reaches the evaluated criteria after approximately 6800--7100 updates, whereas the VB-guided conditions reach them after approximately 1700--2600 updates.
Relative to the baseline, Flat + VB reaches the criteria 2.7--4.0 times earlier ($p\leq0.005$).
Because the conventional baseline and the VB-guided regime may differ in more than the band reward alone, this comparison is interpreted as the effect of the VB-guided training regime rather than as a fully isolated velocity-band effect.

By contrast, adding the tilted-gravity field does not change \emph{when} stepping emerges.
With the reward terms held fixed, Slope + VB and Flat + VB reach every emergence criterion at essentially the same update, and so do Slope + VB + PW and Flat + VB + PW: the paired crossing-update ratios stay within 0.90--1.10 ($p=0.19$--$0.94$), with no consistent advantage for the tilted condition.
This ordering persists after normalizing the step count by episode duration---at update 2000, Flat + VB and Slope + VB produce 0.097 and 0.082 contact events per control step, respectively.

The incremental economy benefit of the tilt therefore cannot be attributed to earlier stepping onset: the initial-tilt sweep below (Table~\ref{tab:tilt_sweep}) shows that adding the tilt lowers mechanical work per meter---and hence CoT---relative to the flat ($0^{\circ}$) condition at an unchanged emergence time.

Because the logged step metric does not distinguish the left and right feet, this analysis concerns sustained stepping acquisition, not gait-cycle periodicity or left--right alternation.

\subsubsection{Initial-Tilt-Angle Sensitivity}
The initial tilt $\theta_{\mathrm{init}}$ is swept over $\{0,3,4,5,6,8\}^{\circ}$.
The $0^{\circ}$ condition coincides with Flat + VB + PW.
Table~\ref{tab:tilt_sweep} reports work per meter over the three common seeds $\{42,123,7\}$.

\begin{table}[!t]
\caption{Initial-Tilt-Angle Sweep (3 Seeds)}
\label{tab:tilt_sweep}
\centering
\begin{tabular}{lccccc}
\hline
$\theta_{\mathrm{init}}$ & $W^{+}/d$ & $(W^{+}{+}W^{-})/d$ & vs.\ $5^{\circ}$ & Surv. & Failed \\
 & [J/m] & [J/m] &  &  & seeds \\
\hline
$0^{\circ}$  & $108.6 \pm 4.2$ & $152.8 \pm 3.1$ & $+4.9$\%  & 0.982 & 0/3 \\
$3^{\circ}$  & $108.1 \pm 0.5$ & $151.6 \pm 1.1$ & $+4.1$\%  & 0.980 & 0/3 \\
$4^{\circ}$  & $104.6 \pm 2.3$ & $149.1 \pm 1.4$ & $+2.4$\%  & 0.975 & 0/3 \\
$\mathbf{5^{\circ}}$ & $\mathbf{102.2 \pm 0.5}$ & $\mathbf{145.6 \pm 1.9}$ & --- & \textbf{0.992} & 0/3 \\
$6^{\circ}$  & $104.8 \pm 1.7$ & $148.9 \pm 3.0$ & $+2.3$\%  & 0.985 & 0/3 \\
$8^{\circ}$  & $124.0 \pm 2.4$ & $199.5 \pm 3.7$ & $+37.0$\% & 0.662 & \textbf{1/3} \\
\hline
\end{tabular}
\end{table}

Work per meter is minimized at $5^{\circ}$, with similar performance over $4^{\circ}$--$6^{\circ}$.
Larger assistance does not yield further improvement: at $8^{\circ}$, one of three seeds fails to converge and survival decreases to 0.662.
A five-seed extension likewise shows two failures at $8^{\circ}$.
These results support interpreting the tilt as a training scaffold with a finite useful range rather than an energy source to be maximized.

\subsubsection{Permanent-Tilt Control}
A separate control evaluates whether the tilt should remain active throughout training.
This condition applies a fixed $5^{\circ}$ simulator-gravity tilt throughout training and is otherwise matched to Flat + VB.
Table~\ref{tab:permanent_tilt} compares this permanent-tilt condition with its no-tilt counterpart.

\begin{table}[!t]
\caption{Permanent $5^{\circ}$ Tilt vs.\ No Tilt, Rewards Identical (Paired Over 3 Seeds: 42, 123, 7)}
\label{tab:permanent_tilt}
\centering
\begin{tabular}{lccccc}
\hline
Speed & CoT & CoT & $\Delta$ & Vel.\ err. & Gait sym. \\
 & (no tilt) & (perm.\ tilt) &  & $\Delta$ & $\Delta$ \\
\hline
0.5 & 0.4512 & 0.4882 & $\mathbf{+8.2\%}$   & $+183.4$\% & $-32.8$\% \\
1.0 & 0.4419 & 0.4554 & $\mathbf{+3.1\%}$   & $+80.7$\%  & $-4.8$\%  \\
1.5 & 0.4779 & 0.5388 & $\mathbf{+12.8\%}$  & $+17.3$\%  & $-3.8$\%  \\
2.0 & 0.5422 & 1.0848 & $\mathbf{+100.1\%}$ & $+161.0$\% & $-8.0$\%  \\
\hline
\end{tabular}
\end{table}

The permanent tilt degrades CoT and tracking at all evaluated speeds.
At 2.0~m/s, CoT approximately doubles and velocity-tracking error increases by 161\%.
Both positive and negative work also increase substantially.
These results indicate that the benefit depends on using the tilt as temporary training assistance and withdrawing it before nominal-dynamics optimization, rather than retaining slope-equivalent energy injection throughout training.

\subsection{Omnidirectional Extension and AMP}
\label{subsec:omni}
We next examine whether the curriculum extends to unassisted omnidirectional locomotion and how it interacts with AMP.
Four conditions form a $2\times2$ comparison over $\{\mathrm{Baseline},\mathrm{Ours}\}$ and $\{\mathrm{no\ AMP},\mathrm{AMP}\}$.
Each condition is trained with three seeds $\{43,42,123\}$ and evaluated over 2000 episodes of randomly sampled commands; we report mean $\pm$ standard deviation across seeds, with contrasts paired over seeds.

Backward commands exposed a bilateral-hopping failure mode during assisted low-speed training.
The bilateral-flight penalty was introduced as a targeted curriculum term and is inactive after the curriculum.
No converged policy exhibits a deployment-time flight gait in the evaluated episodes.

\subsubsection{Efficiency and Stability Under AMP}

Because realized speeds differ across conditions---the AMP policies realize $0.86$--$1.00$~m/s versus about $1.23$~m/s without AMP---Table~\ref{tab:omni} reports speed-matched CoT over the common forward-command speed range of 0.50--1.40~m/s.

\begin{table}[!t]
\footnotesize
\caption{Forward-Command Evaluation under the Omnidirectional Task (Mean $\pm$ S.D., 3 Seeds)}
\label{tab:omni}
\centering
\setlength{\tabcolsep}{3pt}
\begin{tabular}{lccc}
\hline
Condition & CoT & Tracking & Stuck \\
 & (speed-matched) & error [m/s] & rate [\%] \\
\hline
Baseline & $0.5641{\pm}0.0072$ & $0.110{\pm}0.005$ & $0.02{\pm}0.03$ \\
Ours & $0.5431{\pm}0.0196$ & $0.118{\pm}0.009$ & $0.07{\pm}0.12$ \\
Baseline + AMP & $0.4982{\pm}0.0280$ & $0.256{\pm}0.035$ & $15.95{\pm}10.91$ \\
\textbf{Ours + AMP} & $\mathbf{0.4588{\pm}0.0041}$ & $0.244{\pm}0.056$ & $\mathbf{4.60{\pm}4.13}$ \\
\hline
\end{tabular}
\end{table}

\begin{table}[!t]
\footnotesize
\caption{$2\times2$ Decomposition of the CoT Effect (vs.\ Baseline; Mean $\pm$ S.D., 3 Seeds)}
\label{tab:factorial}
\centering
\setlength{\tabcolsep}{3pt}
\begin{tabular}{lcc}
\hline
Effect & Raw [\%] & Speed-matched [\%] \\
\hline
Full PDW-inspired curriculum, no AMP & $-4.70{\pm}2.40$ & $-3.75{\pm}2.68$ \\
Motion prior alone & $-7.99{\pm}9.14$ & $\mathbf{-11.67{\pm}5.30}$ \\
Both & $-17.03{\pm}2.62$ & $-18.66{\pm}1.55$ \\
Additive prediction & $-12.69{\pm}6.75$ & $-15.42{\pm}2.64$ \\
\textbf{Interaction} & $-4.34{\pm}4.13$ & $\mathbf{-3.24{\pm}1.96}$ \\
\hline
\end{tabular}
\end{table}

Table~\ref{tab:factorial} decomposes the speed-matched CoT effect into the curriculum, the motion prior, and their interaction.
Without AMP, the complete curriculum---tilted gravity, the curriculum-coupled objectives, and the synchronized command expansion together---reduces speed-matched CoT by $3.75\pm2.68\%$; the reduction is negative in all three seeds but modest and seed-dependent, and tracking error remains essentially unchanged.
Because this condition is not decomposed further, the effect cannot be attributed to tilted gravity in isolation.
AMP alone reduces speed-matched CoT by $11.67\pm5.30\%$, although the AMP policies walk more slowly and exhibit substantially higher and more variable stuck rates ($15.95\pm10.91\%$ for Baseline + AMP).
The higher stuck rate and tracking error under AMP reflect a longer velocity rise time---the policy accelerates more gradually toward the commanded speed---whereas the no-AMP policy reaches the command faster but overshoots it.
The combined condition yields the largest and most consistent reduction, $18.66\pm1.55\%$, exceeding the additive prediction of the two single effects in every seed (interaction $-3.24\pm1.96\%$, negative in all three seeds).
Equivalently, the motion prior appears to amplify the curriculum's own contribution: the curriculum's marginal CoT reduction is larger in AMP's presence than in its absence ($3.75\%$ to $6.99\%$, the combined effect minus AMP alone), consistent with the two combining more than additively.
This super-additive trend is directionally consistent across seeds but, with three seeds, does not reach statistical significance (one-sample $t=-2.87$, one-sided $p\approx0.05$); we report it as suggestive rather than conclusive, while noting that the hardware evaluation (Section~\ref{subsec:sim2real}) provides independent, converging support---there the curriculum's efficiency advantage is near-neutral on its own but clear once AMP is added.
Combining the curriculum with AMP is also associated with a lower stuck rate ($15.95\%$ to $4.60\%$) and reduced across-seed variance relative to AMP alone, so the curriculum in turn makes the motion prior's efficiency gain more reliable.

\begin{figure*}[!t]
\centering
\includegraphics[width=0.7\textwidth]{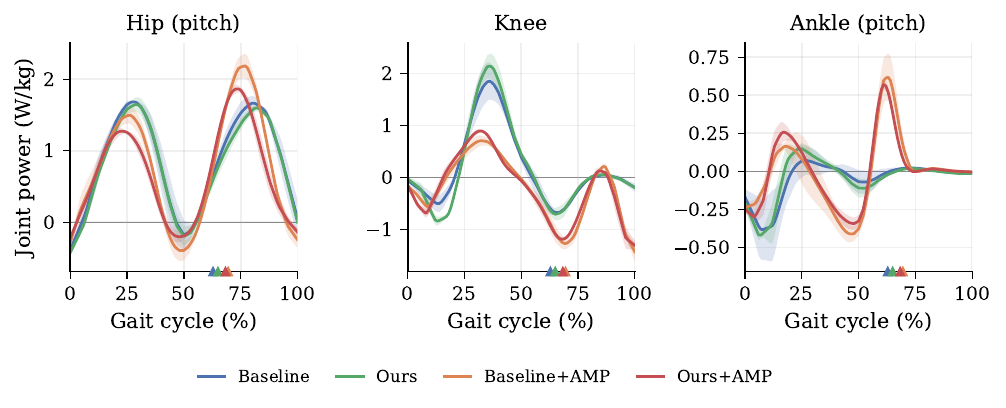}
\caption{Sagittal joint power over the gait cycle, from ipsilateral foot touchdown to the next touchdown of the same foot, for forward locomotion speed-matched over a realized-speed range of 0.50--1.40~m/s. Curves are means across three independently trained seeds and shaded bands denote $\pm$ one standard deviation across seeds; markers on the abscissa indicate each condition's mean toe-off. Power is normalized by the nominal robot mass $m$. Note that the vertical scale differs between panels: the ankle operates at roughly one quarter of the hip and knee power range.}
\label{fig:joint_power_phase}
\end{figure*}

\subsubsection{Phase-Resolved Joint Organization}
To characterize how the motion prior reorganizes locomotion rather than only how much energy it costs, we phase-normalize sagittal joint power from ipsilateral touchdown to the next touchdown of the same foot (Fig.~\ref{fig:joint_power_phase}).
With AMP the ankle develops a positive-power burst in late stance, peaking at $63\pm2\%$ and $61\pm0\%$ of the stride and preceding toe-off in both conditions; $72\pm11\%$ and $62\pm4\%$ of all positive ankle work falls within the final $30\%$ of stance, and ankle push-off work reaches $0.0336\pm0.0065$ and $0.0324\pm0.0020$~J\,kg$^{-1}$ per stride, roughly an order of magnitude larger than the corresponding values without AMP ($0.0045\pm0.0033$ and $0.0021\pm0.0002$~J\,kg$^{-1}$).
Without AMP no comparable burst exists: the ankle peak is far smaller and, for the baseline, not reproducibly located across seeds.
The knee changes role over the same comparison, its net work over the stride turning from positive ($+0.058\pm0.004$ and $+0.061\pm0.012$~J\,kg$^{-1}$) to negative ($-0.105\pm0.022$ and $-0.103\pm0.015$~J\,kg$^{-1}$), while its dominant absorption peak moves from just after touchdown to immediately before it.
Full metrics are given in Table~\ref{tab:appjoint2}.

\subsubsection{Integrated Positive-Work Composition}
Integrated over the stride, the same reorganization appears as a change in composition that tracks the motion prior rather than the curriculum.
Decomposing positive actuator work among the sagittal ankle-pitch, knee, and hip-pitch groups under the same forward, speed-matched evaluation, AMP raises the ankle share from $5.4\%$ to $10.8\%$ in the baseline pair and from $5.1\%$ to $11.3\%$ in the proposed pair, and lowers the knee share from $30.9\%$ to $17.4\%$ and from $34.0\%$ to $20.4\%$, respectively.
The hip share rises correspondingly, from $63.7\%$ to $71.7\%$ and from $60.9\%$ to $68.3\%$, and hip work per stride rises with it, so positive work moves away from the knee toward both the ankle and the hip rather than to the ankle alone.
Curriculum-associated shifts are small relative to these AMP-associated shifts.
Relative to the joint-power composition reported for human walking~\cite{FarrisEtAL2012}, AMP moves the knee contribution toward the reported 14--17\% range, but the robot remains strongly hip-dominant and its ankle share remains far below the reported 40--50\% range.
The full composition is shown in Fig.~\ref{fig:joint_work_share}.

\subsubsection{Directional Extension}
Directional evaluation shows that the curriculum improves lateral robustness under AMP.
The Baseline + AMP policy is poorly tracked and inefficient under pure lateral commands, with high and highly variable CoT across seeds, whereas Ours + AMP improves lateral-command performance, yielding a CoT comparable with that observed for its other moving commands---consistent with the curriculum carrying coordination into non-sagittal commands under the same motion prior.
The omnidirectional results indicate that the framework extends to the full command set; the forward-only motion-prior set, which may underlie the Baseline + AMP lateral difficulty, motivates a more balanced dataset with lateral and backward walking as a direction for further gains.
Detailed per-direction cost-of-transport, velocity-tracking, and gait-symmetry results are provided in Appendix~\ref{app:omni}.

\subsection{Sim-to-Real Evaluation}
\label{subsec:sim2real}
\begin{figure*}[!t]
\centering
\subfloat[Deployment snapshots\label{fig:real_a}]{\includegraphics[width=0.70\textwidth]{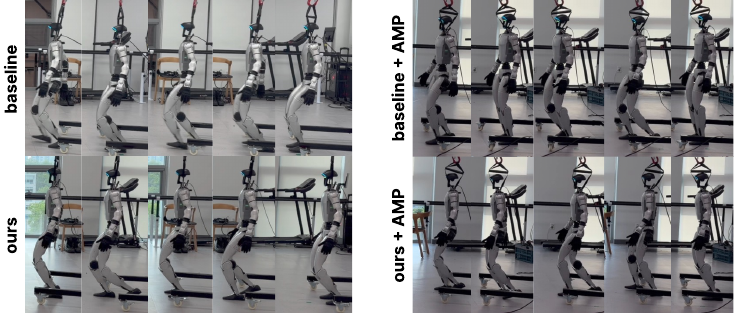}}\hfill
\subfloat[CoT improvement\label{fig:real_b}]{\includegraphics[width=0.27\textwidth]{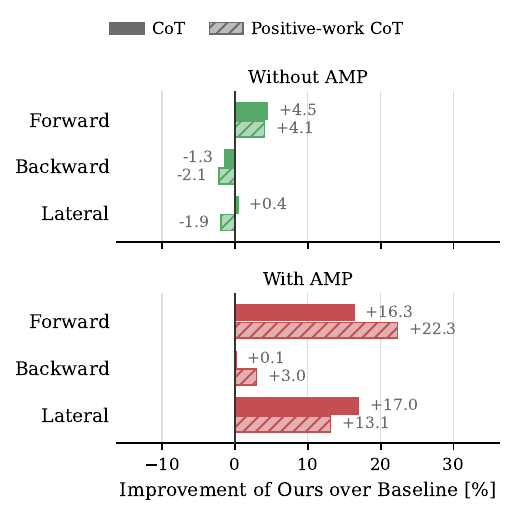}}
\caption{Real-robot deployment on the Unitree G1. (a)~Time-ordered snapshot sequences over roughly one walking stride for the four policies: Baseline and Ours without the motion prior, and Baseline\,+\,AMP and Ours\,+\,AMP. An overhead safety sling is attached in all trials and does not support the robot's weight during walking. (b)~Improvement of Ours over Baseline in cost of transport (solid) and in its positive-work component (hatched) by commanded direction, without AMP (green) and with AMP (red); positive values indicate greater efficiency for Ours, and distances are recovered from onboard sensing (Section~\ref{subsec:hardware}).}
\label{fig:real}
\end{figure*}

We evaluate the learned policies both in simulation under nominal-gravity deployment conditions and on the physical Unitree G1.
Table~\ref{tab:sim2real} reports simulated CoT under randomized and forward commands; Table~\ref{tab:hardware} reports the measured hardware CoT by direction, and Fig.~\ref{fig:real} shows the real-robot deployment together with the direction-level CoT improvement of Ours over Baseline.

\begin{table}[!t]
\footnotesize
\caption{Deployment-Condition CoT in Simulation \\ (Mean $\pm$ S.D., 3 Seeds)}
\label{tab:sim2real}
\centering
\setlength{\tabcolsep}{4pt}
\begin{tabular}{lccc}
\hline
Method & CoT, Random & CoT, Forward & Vel.\ Err. \\
 & Cmd.\ $\downarrow$ & Cmd.\ $\downarrow$ & $\downarrow$ \\
\hline
Baseline & $0.598{\pm}0.025$ & $0.589{\pm}0.009$ & $0.139{\pm}0.004$ \\
Ours & $0.577{\pm}0.024$ & $0.562{\pm}0.021$ & $0.141{\pm}0.005$ \\
$\Delta$ & $+3.5$\% & $+4.7$\% & $-1.7$\% \\
\hline
Baseline + AMP & $0.560{\pm}0.042$ & $0.542{\pm}0.049$ & $0.413{\pm}0.117$ \\
Ours + AMP & $\mathbf{0.481{\pm}0.007}$ & $\mathbf{0.489{\pm}0.012}$ & $\mathbf{0.317{\pm}0.081}$ \\
$\Delta$ & $\mathbf{+14.2\%}$ & $\mathbf{+9.8\%}$ & $\mathbf{+23.2\%}$ \\
\hline
\end{tabular}
\end{table}

In simulation, the curriculum lowers CoT in every deployment condition (Table~\ref{tab:sim2real}): by 3.5--4.7\% without AMP with velocity tracking essentially unchanged, and by up to 14.2\% with AMP, where velocity-tracking error also improves.

\begin{table}[!t]
\footnotesize
\caption{Hardware Cost of Transport by Direction \\ (Measured Distance)}
\label{tab:hardware}
\centering
\setlength{\tabcolsep}{4pt}
\begin{tabular}{llcccc}
\hline
AMP & Direction & Baseline & Ours & $\Delta\mathrm{CoT}$ & $\Delta\mathrm{CoT}^{+}$ \\
\hline
\multirow{3}{*}{No}  & Forward  & 0.581 & 0.555 & $+4.5$\%  & $+4.1$\%  \\
                     & Backward & 0.614 & 0.622 & $-1.3$\%  & $-2.1$\%  \\
                     & Lateral  & 0.587 & 0.585 & $+0.4$\%  & $-1.9$\%  \\
\hline
\multirow{3}{*}{Yes} & Forward  & 0.521 & 0.436 & $\mathbf{+16.3}$\% & $\mathbf{+22.3}$\% \\
                     & Backward & 0.499 & 0.499 & $+0.1$\%  & $+3.0$\%  \\
                     & Lateral  & 0.489 & 0.406 & $\mathbf{+17.0}$\% & $\mathbf{+13.1}$\% \\
\hline
\multicolumn{6}{l}{\footnotesize CoT uses total joint power; $\mathrm{CoT}^{+}$ uses positive work only.} \\
\multicolumn{6}{l}{\footnotesize Positive $\Delta$ indicates greater efficiency for Ours. Lateral pools} \\
\multicolumn{6}{l}{\footnotesize left and right; $n=2$--$6$ steady-state traversals per cell.} \\
\end{tabular}
\end{table}

On hardware (Table~\ref{tab:hardware} and Fig.~\ref{fig:real}), the curriculum's efficiency benefit is strongest and most consistent when it is combined with the motion prior.
Ours\,+\,AMP lowers cost of transport relative to Baseline\,+\,AMP by 16.3\% for forward and 17.0\% for lateral commands, with the positive-work component falling by 22.3\% and 13.1\% respectively, while backward walking is essentially unchanged ($+0.1$\%).
Without the motion prior the hardware CoT differences are small---$+4.5$\% forward, $+0.4$\% lateral, and $-1.3$\% backward---and lie within the trial-to-trial spread; the forward value nonetheless matches the $+4.7$\% forward reduction measured in simulation (Table~\ref{tab:sim2real}).
Each value aggregates $n=2$--$6$ steady-state segments per direction and condition on a single deployed policy per condition, with the travelled distance recovered from the onboard joint kinematics and IMU (Section~\ref{subsec:hardware}) rather than assumed from the commanded velocity.

In the deployment snapshots of Fig.~\ref{fig:real_a}, the two policies within each AMP setting are difficult to tell apart by eye, whereas the two AMP settings differ visibly: without AMP both policies walk with more bent knees, and with AMP both adopt a more upright posture with more natural arm swing. The corresponding walking trials are shown in the video provided as a multimedia attachment.

\section{Discussion}
\label{sec:discussion}

\subsection{Training-Time Dynamics Guidance and Locomotion Economy}
The component ablation of the forward-locomotion study (Section~\ref{subsec:forward}) indicates that reward design and training-time dynamics guidance act as complementary levers for locomotion economy.
Within the velocity-band regime, neither the positive-work penalty nor the tilted-gravity field reproduces the full reduction in cost of transport on its own, and the complete condition improves on both.
This finding aligns with previous studies showing that locomotion economy can be improved through effort penalties or explicit energy constraints in the learning objective~\cite{FuEtAL2021,JeonEtAL2023,HuangEtAL2026}.
The additional CoT reduction observed when dynamics guidance is combined with reward shaping further suggests that the two components improve locomotion economy through complementary mechanisms.
The benefit of dynamics guidance may arise because it changes the conditions under which early policy search takes place, biasing learning toward mechanically efficient coordination rather than acting solely through the objective.
This interpretation is also consistent with previous observations that the balance between energy, tracking, and stability objectives can be sensitive to reward scaling~\cite{JeonEtAL2023}.
Further work could examine whether the additional improvement indeed arises from altered gait exploration during training and whether this benefit generalizes across reward configurations, locomotion tasks, and robot platforms.

What training-time dynamics guidance changes is the economy of the converged gait.
The reduction in cost of transport is concentrated almost entirely in positive work and grows with commanded velocity, while command tracking is preserved (Section~\ref{subsec:work}).
This pattern is consistent with the energetics of human walking, where redirecting the center of mass from one stance leg to the next is a principal determinant of mechanical work and metabolic cost~\cite{DonelanEtAL,KuoEtAL2005}.
It suggests that the lower CoT reflects a reduction in the positive actuator work required for step-to-step coordination rather than energy saved by executing less of the commanded task.
The change is not one of acquisition speed.
The velocity-band objective makes sustained stepping appear much earlier than in the conventional baseline, but the tilted-gravity field does not: with the reward terms held fixed, it leaves the onset of stepping unchanged while still lowering work per meter (Fig.~\ref{fig:gait_emergence}).
In the closest learning-side precedents, temporary assistance mainly makes a target behavior easier to acquire~\cite{RudinEtAL,SiekmannEtAL2021,ShiEtAL2023,CaoEtAL2025}, whereas here the acquisition speedup belongs to the reward terms and the economy gain to the tilted-gravity field.
Temporary dynamics guidance can therefore bias policy search toward mechanically efficient coordination that persists after the guidance is removed.
The present ablation nevertheless varies the positive-work penalty as a whole, without separating its sign from its magnitude; a symmetric penalty of matched magnitude would resolve this.

The tilt-angle sweep and the permanent-tilt control indicate that the tilted-gravity field helps only while it remains bounded and is withdrawn before final adaptation to nominal dynamics.
The response to the initial tilt angle is non-monotonic, with larger tilts degrading convergence, and keeping the tilt active throughout training degrades both economy and tracking at every speed (Section~\ref{subsec:sensitivity}).
A bounded useful range is consistent with the observation in curriculum learning that the largest gains arise from conditions of intermediate difficulty~\cite{FlorensaEtAL2018,NarvekarEtAL2020}, although the quantity varied here is the dynamics under which policy search takes place rather than task difficulty.
This behavior also separates the method from approaches that retain a passive-dynamics mechanism at run time, such as Virtual Slope Walking~\cite{DongEtAL2011} and the Duke Humanoid~\cite{XiaEtAL2025}, where the mechanism remains part of the deployed system; a numerical comparison is not attempted, since the platforms and cost definitions differ.
The tilt is therefore best understood as a property of the training procedure and not of the deployed system.

\subsection{Generalization and Interaction with the Motion Prior}
The benefit carries over to unassisted omnidirectional locomotion, but its size depends on the command direction, with the largest reduction on forward commands and smaller reductions elsewhere (Appendix~\ref{app:omni}).
This dependence follows from the geometry of the tilted-gravity field, which is defined in the sagittal plane and has no analogue for lateral progression.
It is also consistent with human data, in which lateral balance is actively regulated rather than passively maintained~\cite{BaubyEtAL2000} and sideways walking is metabolically expensive relative to forward walking at comparable speeds~\cite{HandfordEtAL2014}.
The motion prior carries a similar bias for a different reason, since its reference set contains forward walking only.
Defining the tilt in the frontal plane and adding lateral and backward walking to the reference set would test whether this direction dependence shifts with these two design choices.

The motion prior also introduces a reliability trade-off that the curriculum offsets.
Added on its own, the prior improves mechanical economy, consistent with reports that adversarial motion priors yield energy-efficient gaits~\cite{EscontrelaEtAL2022}.
It also raises the stuck rate and the tracking error, however, and its gain is strongly seed-dependent, plausibly reflecting both a longer velocity rise time and the training instability of adversarial objectives~\cite{TangEtAL2024,VollenweiderEtAL2023}.
Adding the curriculum recovers what the prior costs: it lowers the stuck rate, reduces the across-seed variance, restores lateral performance, and yields the lowest CoT of the four conditions, with a combined reduction that exceeds the additive prediction in every seed (Section~\ref{subsec:omni}).
A plausible reason is that the velocity band withholds reward for stationary or reversed motion during early policy search, which is the same failure to make progress that the prior-only policies show in their elevated stuck rate.
With three seeds the interaction itself falls just short of significance ($p\approx0.05$), so the claim is narrow: the curriculum remains effective in the presence of a walking-specific motion prior, and it makes the efficiency gain of that prior more reliable.

Joint-level organization is governed by a different part of the framework than mechanical economy.
With the motion prior the ankle share of positive work roughly doubles, the knee share falls, and the hip share rises correspondingly, so positive work moves away from the knee toward both the ankle and the hip.
The knee also turns from a net generator over the stride into a net absorber.
The curriculum, by contrast, leaves the sagittal work composition and the phase-resolved profiles nearly unchanged (Section~\ref{subsec:omni}).
This division follows from what each component specifies: the prior supplies reference-derived temporal and distal-joint structure~\cite{PengEtAL2021,PengEtAL2018}, while the curriculum changes the conditions of early policy search without prescribing distal-joint organization.
The reorganization is a byproduct rather than a target, since the prior never evaluates actuator work, and it does not explain the CoT reduction, because without the prior the curriculum lowers CoT while slightly lowering the ankle share.
The resulting organization also matches human walking only in part: the knee contribution approaches the range reported for human walking, whereas the ankle stays far below its reported share and the robot remains strongly hip-dominant~\cite{FarrisEtAL2012}.
Part of this gap is a difference in the measured quantity, since the human values are inverse-dynamics powers that include elastic tendon contributions~\cite{FukunagaEtAL2001} whereas ours are actuator mechanical work; a passive ankle spring alone lowers metabolic cost by about 7\% without supplying net positive work~\cite{CollinsEtAL2015}.
A kinematically scored objective thus transfers the reference body's kinematics but not its energetic organization.

\subsection{Sim-to-Real Relevance}
The hardware results are consistent with simulation in the sagittal plane, and the agreement is clearest under the motion prior (Section~\ref{subsec:sim2real}).
The hardware reproduces the direction of the simulated effect and the dependence on the prior seen in the factorial decomposition, where the marginal benefit of the curriculum is larger in the prior's presence.
The curriculum-only differences nevertheless lie within the trial-to-trial spread, so the close numerical agreement between hardware and simulation is best read as consistency rather than as an independent confirmation of effect size.
Backward locomotion is the exception: in simulation it shows the largest reduction of any direction when the motion prior is present, whereas on hardware it is essentially unchanged with the motion prior and slightly adverse without it (Table~\ref{tab:hardware}).
Backward walking is not a simple reversal of forward walking, since joint kinematics are close to time-reversed but joint powers are nearly reversed in polarity~\cite{WinterEtAL1989} and the underlying muscle synergies are organized differently~\cite{GrassoEtAL1998}.
Backward commands do receive assistance, but at a reduced magnitude ($\theta_{\mathrm{back}}=3^{\circ}$ against $5^{\circ}$ forward), so the mechanism that lowers forward positive work need not act in the same way under an inverted power profile.
Backward is also the direction with the fewest evaluation episodes in simulation and the only direction that required a dedicated curriculum term during assisted training, and we therefore do not treat the simulated backward benefit as validated on hardware.

\subsection{Limitations and Future Work}
Several limitations bound the scope of these conclusions.
All results are obtained on a single humanoid platform, so generality across morphologies and actuator architectures remains untested, and the tilt schedule and the initial tilt angle are fixed by design rather than adapted online, so both would need to be re-identified on a different robot.
The two simulation studies also differ in ablation scope: the forward study decomposes the framework into tilted gravity, the velocity band, and the positive-work penalty across five seeds, whereas the omnidirectional condition is evaluated as a whole across three seeds, so component-level attribution rests on the forward study.
The hardware evaluation is likewise limited, since it uses a single deployed policy per condition and its kinematic distance recovery assumes flat-ground, no-slip stance contact, which was not independently measured on the physical robot.

Further limitations concern what the reported measures cover.
The mechanical CoT based on $\sum_j|\tau_j\dot q_j|$ measures actuator mechanical work only, excluding Joule heating, transmission losses, and driver power.
It also does not distinguish a gait that covers ground with long strides from one that shuffles with short, rapid steps at the same average speed; we make no claim of orbital stability or of a passive limit cycle for the learned policies.
Because CoT is evaluated on the episodes in which a policy made progress, conditions differing in stuck rate are compared on slightly different episode sets, which is negligible without the motion prior and favors the baseline with it.
The increase in negative knee work is also interpreted as joint-level energy absorption rather than as reduced impact loading, which was not measured, and the phase-resolved evaluation uses forward-only commands, so it is matched in realized speed but not in command distribution with the work-share and cost-of-transport evaluations.

\section{Conclusion}
\label{sec:conclusion}
We investigated passive-dynamics-inspired assistance as a temporary condition for humanoid locomotion learning rather than as a mechanism retained during deployment.
After the tilted-gravity field and curriculum-coupled rewards were removed, the resulting policies maintained nominal flat-ground command tracking at a lower mechanical cost, with the reduction concentrated primarily in positive actuator work.
With the reward terms held fixed, the tilted-gravity field acted on which gait was retained rather than on how quickly stepping emerged.
The sensitivity and permanent-assistance controls further bound where it helps: too large an initial tilt, or a tilt that is never withdrawn, degraded both economy and tracking.

The benefit persisted when a walking-specific motion prior was added, with a consistent greater-than-additive trend that the present evidence leaves suggestive.
A preliminary physical-robot evaluation reproduced the simulated forward and lateral improvements once the motion prior was present, although the curriculum-only differences fall within the trial-to-trial spread and backward locomotion did not reproduce its simulated benefit.

More broadly, these findings suggest that energy-efficient locomotion can be shaped not only through the objective optimized by a policy, but also through the physical conditions under which the policy discovers its gait.
A natural direction for future work is to use the learned joint-power profiles to guide the placement and tuning of passive elastic or damping elements in humanoid hardware.
Such morphology--policy co-design could further reduce actuator demand and would complement passive-dynamics-oriented hardware designs~\cite{XiaEtAL2025}.

\appendices

\section{Implementation Details}
\label{app:impl}
All conditions share the simulation, actuation, network, optimization, and
domain-randomization settings of Tables~\ref{tab:impl_sim}
and~\ref{tab:impl_ppo}, and the baseline locomotion reward
$r^{\mathrm{base}}$ of Table~\ref{tab:impl_basereward}.
They differ only in the training-time components under study, listed in
Tables~\ref{tab:impl_pdw} and~\ref{tab:impl_amp}.

\begin{table}[!h]
\footnotesize
\caption{Simulation, Actuation, and Domain Randomization}
\label{tab:impl_sim}
\centering
\begin{tabular}{ll}
\hline
Item & Value \\
\hline
\multicolumn{2}{l}{\textit{Simulation and control}} \\
Simulator & Isaac Lab 2.3.0 / Isaac Sim 5.1.0 (PhysX) \\
Physics step / decimation & 0.005~s / 4 \\
Control period $\Delta t$ & 0.02~s (50~Hz) \\
Episode length & 20~s (1000 control steps) \\
Parallel environments & 1024 \\
Robot / nominal model mass & Unitree G1, 29 DoF / 33.34~kg \\
Deployment gravity & $(0,0,-9.81)$~m/s$^2$ \\
\hline
\multicolumn{2}{l}{\textit{Actuation (implicit PD, $K_p$ / $K_d$)}} \\
Hip pitch / roll / yaw & 100 / 2.0 \\
Knee & 150 / 4.0 \\
Ankle & 40 / 2.0 \\
Waist yaw & 200 / 5.0 \\
Waist roll / pitch & 40 / 5.0 \\
Arm joints & 40 / 1.0 \\
Action scale / armature & 0.25 / 0.01 \\
\hline
\multicolumn{2}{l}{\textit{Domain randomization (identical in every condition)}} \\
Friction (per body) & sampled $[0.3, 1.0]$ \\
Torso added mass & $[-1.0, +3.0]$~kg about nominal \\
Push perturbation & base vel.\ $\pm0.5$~m/s in $x,y$ every 5~s \\
Reset base pose & $\pm0.5$~m in $x,y$; uniform yaw \\
Reset joint velocity & $\pm1.0$~rad/s \\
\hline
\end{tabular}
\end{table}

\begin{table}[!h]
\footnotesize
\caption{Policy / Critic Network and PPO Hyperparameters}
\label{tab:impl_ppo}
\centering
\begin{tabular}{ll}
\hline
Item & Value \\
\hline
Recurrent core & single-layer LSTM, 256 units \\
MLP head hidden sizes & $[256, 128]$ \\
Observation stack & last 5 control steps \\
Actor input & $96 \times 5 = 480$ \\
Critic input & $99 \times 5 = 495$ (adds base lin.\ vel.) \\
Learning rate & $10^{-3}$ (adaptive), target KL 0.01 \\
Discount $\gamma$ / GAE $\lambda$ & 0.99 / 0.95 \\
Clip $\epsilon$ / entropy coeff. & 0.2 / 0.01 \\
Value-loss coeff. / grad clip & 1.0 / 1.0 \\
Epochs / minibatches per update & 5 / 4 \\
Steps per env per update & 24 (24{,}576 transitions) \\
Total updates & 15{,}000 ($\approx$368.6~M env steps) \\
\hline
\end{tabular}
\end{table}

The reward manager forms the per-step return as
$r_t = \Delta t \sum_i w_i\, f_i(s_t,a_t)$ with $\Delta t = 0.02$~s, so the
weights $w_i$ in Table~\ref{tab:impl_basereward} are the raw coefficients
before this $\Delta t$ scaling.

\begin{table*}[!t]
\footnotesize
\caption{Baseline Locomotion Reward $r^{\mathrm{base}}$, Identical Across All Conditions}
\label{tab:impl_basereward}
\centering
\begin{tabular}{cllr}
\hline
\# & Term & Formula $f_i$ & Weight $w_i$ \\
\hline
1 & Linear-velocity tracking & $\exp\!\big(-\lVert \mathbf{c}_{v}-\mathbf{v}^{b}_{xy}\rVert^2/\sigma^2\big)$ & $+1.0$ \\
2 & Angular-velocity tracking (yaw) & $\exp\!\big(-(\omega_z^{*}-\omega_{b,z})^2/\sigma^2\big)$ & $+1.0$ \\
3 & Alive bonus & $\mathbf{1}[\,\text{not terminated}\,]$ & $+0.15$ \\
4 & Base angular velocity (roll/pitch) & $\lVert \boldsymbol{\omega}_{b,xy}\rVert^2$ & $-0.05$ \\
5 & Base orientation & $\lVert \mathbf{g}_{b,xy}\rVert^2$ & $-1.0$ \\
6 & Joint velocity & $\sum_j \dot{q}_j^2$ & $-1{\times}10^{-3}$ \\
7 & Joint acceleration & $\sum_j \ddot{q}_j^2$ & $-2.5{\times}10^{-7}$ \\
8 & Action rate & $\kappa(\mathbf{c})\,\lVert \mathbf{a}_t-\mathbf{a}_{t-1}\rVert^2$ & $-0.05$ \\
9 & Joint-limit violation & $\sum_j\big([q_j^{\min}-q_j]_+ + [q_j-q_j^{\max}]_+\big)$ & $-5.0$ \\
10 & Energy & $\sum_{j=1}^{29} |\tau_j|\,|\dot{q}_j|$ & $-2{\times}10^{-5}$ \\
11 & Arm posture deviation & $\kappa(\mathbf{c})\sum_{j\in\text{arms}} |q_j-\bar{q}_j|$ & $-0.1$ \\
12 & Waist posture deviation & $\kappa(\mathbf{c})\sum_{j\in\text{waist}} |q_j-\bar{q}_j|$ & $-1.0$ \\
13 & Leg posture deviation (hip roll/yaw) & $\kappa(\mathbf{c})\sum_{j\in\text{hip}} |q_j-\bar{q}_j|$ & $-1.0$ \\
14 & Undesired contacts & $\sum_b \mathbf{1}[\lVert \mathbf{F}_b\rVert > 1~\text{N}]$ & $-1.0$ \\
\hline
\end{tabular}

\vspace{3pt}
\begin{minipage}{0.92\textwidth}
\footnotesize
Symbols are as defined in Section~\ref{sec:method}; $\bar q$ is the default pose and $\mathbf{F}_b$ the contact force on body $b$. Tracking kernels (terms 1--2) use $\sigma=0.5$. Stand-still boost: $\kappa(\mathbf{c})=2.0$ when $\lVert\mathbf{c}_{v}\rVert<0.1$~m/s and $|\omega_z^{*}|<0.05$~rad/s, else $\kappa(\mathbf{c})=1.0$; it sharpens terms 8 and 11--13 under a near-zero command to suppress standing jitter. Term 14 sums over all bodies except the ankles. Joint groups---arms: shoulder, elbow, wrist; waist: waist joints; legs (posture term only): hip roll and hip yaw; terms 6, 7, 9, and 10 act on all 29 DoF.
\end{minipage}
\end{table*}

\begin{table}[!h]
\footnotesize
\caption{PDW-Inspired Reward Terms and Curriculum Schedule}
\label{tab:impl_pdw}
\centering
\setlength{\tabcolsep}{3pt}
\begin{tabular}{lcl}
\hline
Term / parameter & Value & Gate / meaning \\
\hline
\multicolumn{3}{l}{\textit{Reward terms, each multiplied by $\sigma(k)$}} \\
Positive-work penalty & $5\times10^{-4}$ & slope-gated \\
Velocity-band reward & 0.3 & slope-gated \\
\quad lower threshold $v_{\min}$ & --- & $0.05$~m/s \\
\quad upper bound (forward study) & --- & $1.0$~m/s, absolute \\
\quad upper bound (omnidir.) & $\eta_{\mathrm{VB}}$ & $1.2\,s_t^{*}$ \\
\quad cmd.\ speed (forward study) & --- & $(0,0.5)\!\to\!(0,2.0)$~m/s \\
\quad cmd.\ speed (omnidir.) & --- & Section~\ref{sec:setup} \\
Bilateral-flight penalty & 1.0 & $v_{\mathrm{thr}}{=}1.0$~m/s, slope-gated \\
\hline
\multicolumn{3}{l}{\textit{Curriculum schedule}} \\
$\theta_{\mathrm{init}}$ & $5^{\circ}$ & forward initial tilt \\
$\theta_{\mathrm{back}}$ & $3^{\circ}$ & maximum backward tilt \\
$K_w$ (omnidirectional) & 150 ep & end of full assistance \\
$K_w$ (forward study) & 100 ep & end of full assistance \\
$K_t$ & 200 ep & end of linear transition \\
$K_{\mathrm{final}}$ & 360 ep & full command range \\
\hline
\end{tabular}

\vspace{2pt}
\begin{minipage}{0.95\columnwidth}
\footnotesize
Episode counts map to PPO updates as $150$, $100$, $200$, and $360$ episodes $\approx 6250$, $4167$, $8333$, and $15{,}000$ updates. With episode counter $k$, $\beta(k)=\mathrm{clip}((k-K_w)/(K_t-K_w),0,1)$, $\sigma(k)=1-\beta(k)$, and $\theta(k)=\theta_{\mathrm{init}}\,\sigma(k)$; roughly $44\%$ of training proceeds after the tilt is withdrawn. The forward commanded-speed range expands by $\pm0.1$~m/s per qualifying episode (tracking threshold 0.6).
\end{minipage}
\end{table}

The adversarial motion prior of Section~\ref{subsec:amp} uses 127 walking clips
from AMASS retargeted to the G1's 29 DoF and resampled to 50~Hz
(42{,}398 state transitions).

\begin{table}[!h]
\footnotesize
\caption{AMP Motion Feature $\Phi$ (80-Dim) and Discriminator}
\label{tab:impl_amp}
\centering
\begin{tabular}{ll}
\hline
Item & Value \\
\hline
Projected gravity / root height & 3 / 1 \\
Root lin.\ / ang.\ velocity (yaw frame) & 3 / 3 \\
Joint positions / velocities & 29 / 29 \\
Pelvis-relative key-body positions & 12 \\
Feature total & 80 \\
\hline
Discriminator MLP & $160$--$1024$--$512$--$1$ \\
Objective & least-squares GAN \\
Gradient penalty ($\lambda$) / logit reg. & 10 / 0.05 \\
Optimizer / replay buffer & Adam $10^{-4}$ / $10^{5}$ \\
Batch (real + policy) & $4096 + 4096$ \\
Combined reward & $0.9\,r^{\mathrm{task}} + 0.1\,r^{\mathrm{AMP}}$ \\
\hline
\end{tabular}
\end{table}

\section{Forward-Locomotion Study: Detailed Results}
\label{app:forward}
Values use the nominal model mass $m=33.34$~kg, consistent with the main text; percentage reductions are invariant to this choice.
For the emergence analysis of Section~\ref{subsec:sensitivity}, logged statistics are smoothed with a 500-update moving average, and the supporting event-count criteria are the first update at which the mean number of step events per episode exceeds 25 and 50.

\begin{table}[!h]
\footnotesize
\caption{Cost of Transport, Sagittal Task \\ (Mean $\pm$ S.D., 5 Seeds)}
\label{tab:appd1}
\centering
\setlength{\tabcolsep}{3.5pt}
\begin{tabular}{lcccc}
\hline
Condition & 0.5~m/s & 1.0~m/s & 1.5~m/s & 2.0~m/s \\
\hline
Flat (baseline) & $0.4463$ & $0.4487$ & $0.5017$ & $0.5791$ \\
Flat + VB & $0.4508$ & $0.4491$ & $0.4861$ & $0.5579$ \\
Flat + VB + PW & $0.4260$ & $0.4222^{*}$ & $0.4553^{*}$ & $0.5103^{**}$ \\
Slope + VB & $0.4404$ & $0.4392$ & $0.4765$ & $0.5339^{*}$ \\
\textbf{Slope + VB + PW} & $\mathbf{0.4161}^{***}$ & $\mathbf{0.4051}^{***}$ & $\mathbf{0.4356}^{**}$ & $\mathbf{0.4910}^{***}$ \\
\textit{Full vs.\ baseline} & \textit{$-6.8\%$} & \textit{$-9.7\%$} & \textit{$-13.2\%$} & \textit{$-15.2\%$} \\
\hline
\multicolumn{5}{l}{\footnotesize $^{*}p<0.05$, $^{**}p<0.01$, $^{***}p<0.001$ (paired $t$-test vs.\ baseline).} \\
\end{tabular}
\end{table}

Velocity-tracking error is statistically indistinguishable across all five conditions (e.g.\ at 1.0~m/s, baseline $0.0850\pm0.0053$ vs.\ full $0.0881\pm0.0097$~m/s), gait symmetry stays within 0.93--0.98, and push-recovery survival is saturated (0.982--0.988).

\section{Omnidirectional and Directional Evaluation}
\label{app:omni}
All values are means $\pm$ standard deviation across the three seeds $\{43,42,123\}$. Random-command episodes are binned by the commanded direction; Spin and stand commands ($|\mathbf{v}_{\mathrm{cmd}}|<0.3$~m/s), which account for 122 of the 2000 episodes per seed, are excluded because transport-normalized cost becomes ill-conditioned as $d\to0$. Table~\ref{tab:appf1} reports CoT by commanded direction, and Table~\ref{tab:appf2} the corresponding reductions.

\begin{table}[!ht]
\scriptsize
\setlength{\tabcolsep}{1.5pt}
\caption{CoT by Commanded Direction (Mean $\pm$ S.D., 3 Seeds; $n$ = per-seed episodes per bin)}
\label{tab:appf1}
\centering
\begin{tabular}{lcccc}
\hline
Bin ($n$) & Baseline & Ours & Baseline + AMP & Ours + AMP \\
\hline
forward (368) & $0.573{\pm}0.009$ & $0.544{\pm}0.022$ & $0.510{\pm}0.038$ & $\mathbf{0.470{\pm}0.008}$ \\
backward (55) & $0.544{\pm}0.026$ & $0.530{\pm}0.013$ & $0.555{\pm}0.064$ & $\mathbf{0.441{\pm}0.016}$ \\
lateral (272) & $0.529{\pm}0.030$ & $0.514{\pm}0.020$ & $0.525{\pm}0.040$ & $\mathbf{0.435{\pm}0.011}$ \\
walk\_turn (1183) & $0.581{\pm}0.016$ & $0.561{\pm}0.018$ & $0.562{\pm}0.047$ & $\mathbf{0.479{\pm}0.007}$ \\
\textbf{all moving (1878)} & $0.571{\pm}0.017$ & $0.550{\pm}0.019$ & $0.546{\pm}0.042$ & $\mathbf{0.470{\pm}0.006}$ \\
\hline
\end{tabular}
\end{table}

\begin{table}[!h]
\footnotesize
\caption{Directional CoT Reduction (\%)}
\label{tab:appf2}
\centering
\begin{tabular}{lcc}
\hline
Bin & Ours vs.\ Baseline & Ours + AMP vs.\ Baseline + AMP \\
\hline
forward & 5.2 & 8.0 \\
backward & 2.5 & 20.6 \\
lateral & 2.9 & 17.1 \\
walk\_turn & 3.3 & 14.7 \\
\textbf{all moving} & \textbf{3.6} & \textbf{14.0} \\
\hline
\end{tabular}
\end{table}

Without AMP the benefit is largest for forward commands (5.2\%) and smaller in the other directions (2.5--3.3\%), consistent with the sagittal-plane mechanics of the assistance; with AMP the reduction is substantially larger in every direction (8.0--20.6\%) and is no longer concentrated in the sagittal plane. Across the fixed-command directional set, mean velocity-tracking error is 0.115, 0.113, 0.256, and 0.205~m/s and mean gait symmetry is 0.849, 0.870, 0.823, and 0.817 for Baseline, Ours, Baseline + AMP, and Ours + AMP, respectively. Baseline + AMP remains poorly tracked in the pure lateral commands, with high and highly variable CoT across seeds, whereas Ours + AMP improves lateral-command performance, yielding a CoT comparable with that observed for its other moving commands (0.41--0.42).

\section{Joint-Level Work and Phase-Resolved Power}
\label{app:joint}
All values are means $\pm$ sample standard deviations across the three independently trained seeds $\{43,42,123\}$, computed under the protocol of Section~\ref{subsubsec:jointmetrics}.

Strides spanning an episode reset, or with duration outside $0.20$--$1.5$~s or single-leg stance fraction outside $0.30$--$0.95$, are discarded, and toe-off is taken as the first non-contact sample.
Accepted strides are resampled to 101 phase points, so phase is resolved to $1\%$ of the stride, and power is normalized by the nominal robot mass $m$; strides are binned by mean speed in $0.1$~m/s steps over $0.5$--$1.4$~m/s and averaged with equal weight, with all nine bins populated in every condition and seed.
Per-stride quantities are computed for each stride individually, as its phase-normalized power integrated over the window scaled by that stride's own duration, and then averaged over strides, within speed bins, across bins, and across seeds, rather than by scaling a seed-mean power curve by the mean stride duration.

\begin{figure}[!h]
\centering
\includegraphics[width=\columnwidth]{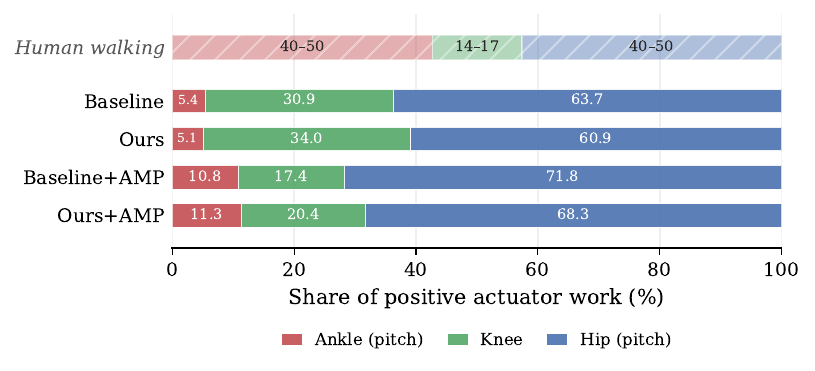}
\caption{Distribution of positive actuator work among the sagittal leg joints during forward locomotion, speed-matched over a realized-speed range of 0.50--1.40~m/s. Each robot bar sums to 100\% within the three sagittal joint groups (Section~\ref{subsubsec:jointmetrics}). Bars are means over three seeds, and the across-seed standard deviation is at most 2.9 percentage points for every segment. The hatched row labels the walking ranges reported by Farris and Sawicki~\cite{FarrisEtAL2012}, with segment widths set by the normalized midpoints of those ranges; those three ranges are reported independently and need not sum to $100\%$, so the row is a set of reference intervals rather than a partition.}
\label{fig:joint_work_share}
\end{figure}

The work-share analysis of Fig.~\ref{fig:joint_work_share} is normalized within the three sagittal joint groups, which together account for 76--79\% of the positive work over all 29 actuators across the four conditions.

\begin{table}[!h]
\footnotesize
\caption{Phase-Resolved Sagittal Joint-Power Metrics \\ (Mean $\pm$ S.D., 3 Seeds)}
\label{tab:appjoint2}
\centering
\setlength{\tabcolsep}{2pt}
\begin{tabular}{lcccc}
\hline
Metric & Baseline & Ours & \begin{tabular}{@{}c@{}}Baseline\\+ AMP\end{tabular} & \begin{tabular}{@{}c@{}}Ours\\+ AMP\end{tabular} \\
\hline
Ankle peak [W/kg] & $0.10{\pm}0.08$ & $0.17{\pm}0.02$ & $0.64{\pm}0.14$ & $0.57{\pm}0.03$ \\
Peak phase [\%] & $40{\pm}29$ & $23{\pm}4$ & $63{\pm}2$ & $61{\pm}0$ \\
Toe-off [\%] & $62.8{\pm}3.6$ & $65.0{\pm}0.4$ & $69.5{\pm}0.4$ & $68.3{\pm}1.0$ \\
Push-off frac.\ [\%] & $14{\pm}19$ & $0{\pm}0$ & $72{\pm}11$ & $62{\pm}4$ \\
$W^{+}_{\mathrm{ank,PO}}$ & $0.0045$ & $0.0021$ & $0.0336$ & $0.0324$ \\
 & ${\pm}0.0033$ & ${\pm}0.0002$ & ${\pm}0.0065$ & ${\pm}0.0020$ \\
$W_{\mathrm{knee}}$ & $+0.058$ & $+0.061$ & $-0.105$ & $-0.103$ \\
 & ${\pm}0.004$ & ${\pm}0.012$ & ${\pm}0.022$ & ${\pm}0.015$ \\
$W_{\mathrm{hip}}$ & $+0.273$ & $+0.280$ & $+0.361$ & $+0.344$ \\
 & ${\pm}0.015$ & ${\pm}0.013$ & ${\pm}0.015$ & ${\pm}0.011$ \\
$W_{\mathrm{knee,WA}}$ & $-0.0052$ & $-0.0063$ & $-0.0170$ & $-0.0235$ \\
 & ${\pm}0.0019$ & ${\pm}0.0025$ & ${\pm}0.0026$ & ${\pm}0.0032$ \\
TD-burst peak [W/kg] & $-0.51$ & $-0.84$ & $-1.35$ & $-1.27$ \\
 & ${\pm}0.17$ & ${\pm}0.09$ & ${\pm}0.15$ & ${\pm}0.12$ \\
TD-burst phase [\%] & $13.7{\pm}0.6$ & $13.0{\pm}0.0$ & $99{\pm}0$ & $99{\pm}0$ \\
Stride period [s] & $0.313$ & $0.344$ & $0.469$ & $0.500$ \\
 & ${\pm}0.022$ & ${\pm}0.003$ & ${\pm}0.005$ & ${\pm}0.024$ \\
\hline
\multicolumn{5}{l}{\footnotesize Work terms are signed and given in J\,kg$^{-1}$ per stride. $W^{+}_{\mathrm{ank,PO}}$ is} \\
\multicolumn{5}{l}{\footnotesize positive ankle work in the final 30\% of stance; $W_{\mathrm{knee,WA}}$ is knee work in} \\
\multicolumn{5}{l}{\footnotesize the first 15\% of stance; $W_{\mathrm{knee}}$ and $W_{\mathrm{hip}}$ are net work over the full stride.} \\
\multicolumn{5}{l}{\footnotesize The TD burst is the knee absorption peak in the $\pm20\%$ interval spanning} \\
\multicolumn{5}{l}{\footnotesize touchdown. Phases are resolved to 1\% of the stride.} \\
\end{tabular}
\end{table}

Quantities integrated over a phase window depend on where that window is placed, which is why the main text reports net work over the full stride.
Knee net work over late stance illustrates this: it is clearly positive without AMP under a final-40\% window but crosses zero under the final-30\% window used here, whereas the full-stride net separates the conditions without overlap.
Ankle net work over the final 30\% of stance behaves similarly, at $-0.0024\pm0.0048$ and $-0.0053\pm0.0022$~J\,kg$^{-1}$ without AMP against $+0.0204\pm0.0078$ and $+0.0193\pm0.0032$~J\,kg$^{-1}$ with it; unlike the positive push-off component, this is a difference of positive and negative contributions inside a window of chosen extent, and is reported for completeness rather than as a primary result.
Knee negative work over swing grows from $-0.0238\pm0.0079$ and $-0.0225\pm0.0036$ to $-0.0872\pm0.0080$ and $-0.0978\pm0.0153$~J\,kg$^{-1}$, and whole-stride ankle net work moves from $-0.016\pm0.011$ and $-0.016\pm0.003$~J\,kg$^{-1}$, a net absorber, to $-0.002\pm0.004$ and $+0.003\pm0.010$~J\,kg$^{-1}$, indistinguishable from zero.

\section{Policy-Architecture Check: Recurrent vs.\ Feed-Forward}
\label{app:arch}
Section~\ref{subsec:policy} fixes the policy to a single-layer LSTM but does not test whether the recurrence is required. We retrain the two core frameworks---Baseline and Ours---with the LSTM replaced by a feed-forward MLP actor and critic (hidden sizes $[512,256,128]$, ELU activations), holding everything else identical: observation stack, PDW-inspired rewards and curriculum, PPO hyperparameters, 1024 environments, 15{,}000 updates, and three seeds $\{43,42,123\}$. Evaluation follows the random-command protocol of Appendix~\ref{app:omni}.

\begin{table}[!h]
\footnotesize
\caption{Random-Command Evaluation: Feed-Forward (MLP) vs.\ Recurrent (LSTM) Policy}
\label{tab:apparch}
\centering
\setlength{\tabcolsep}{3pt}
\begin{tabular}{llccc}
\hline
Framework & Policy & CoT & Vel.\ error [m/s] & Stuck rate \\
\hline
Baseline & MLP & $0.620{\pm}0.018$ & $0.169{\pm}0.002$ & $0.002{\pm}0.001$ \\
Baseline & LSTM & $0.580$ & $0.142$ & $0.001$ \\
Ours & MLP & $0.576{\pm}0.005$ & $0.154{\pm}0.009$ & $0.001{\pm}0.000$ \\
Ours & LSTM & $0.561$ & $0.143$ & $0.000$ \\
\hline
\multicolumn{5}{l}{\footnotesize CoT is the raw overall-command value at $m{=}33.34$~kg (Table~\ref{tab:impl_sim}),} \\
\multicolumn{5}{l}{\footnotesize a few percent above the re-binned all-moving CoT of Table~\ref{tab:appf1}.} \\
\multicolumn{5}{l}{\footnotesize MLP rows are 3-seed means $\pm$ s.d.; LSTM rows reproduce the} \\
\multicolumn{5}{l}{\footnotesize main-study policy for reference.} \\
\end{tabular}
\end{table}

The feed-forward policy reproduces the recurrent one on the core task: CoT agrees to within $3$--$7\%$, velocity tracking is comparable (below $0.17$~m/s error, with stuck rate below $0.3\%$), and the improvement of Ours over Baseline is preserved under both architectures ($0.576<0.620$ for the MLP and $0.561<0.580$ for the LSTM). The PDW-inspired dynamics guidance is thus a property of the training scheme rather than of the network. We nonetheless retain the LSTM because the AMP-augmented variant requires it: with AMP enabled (Section~\ref{subsec:amp}), the same MLP fails to reach functional locomotion within the matched $15{,}000$-update budget (random-command stuck rate above $85\%$, versus below $20\%$ for the LSTM), consistent with AMP acting as a temporal coordination objective that benefits from the recurrent core. The architecture comparison above is therefore restricted to the core, AMP-free frameworks.

\bibliographystyle{IEEEtran}
\bibliography{references}

\begin{thebibliography}{10}
\providecommand{\url}[1]{#1}
\csname url@samestyle\endcsname
\providecommand{\newblock}{\relax}
\providecommand{\bibinfo}[2]{#2}
\providecommand{\BIBentrySTDinterwordspacing}{\spaceskip=0pt\relax}
\providecommand{\BIBentryALTinterwordstretchfactor}{4}
\providecommand{\BIBentryALTinterwordspacing}{\spaceskip=\fontdimen2\font plus
\BIBentryALTinterwordstretchfactor\fontdimen3\font minus
  \fontdimen4\font\relax}
\providecommand{\BIBforeignlanguage}[2]{{%
\expandafter\ifx\csname l@#1\endcsname\relax
\typeout{** WARNING: IEEEtran.bst: No hyphenation pattern has been}%
\typeout{** loaded for the language `#1'. Using the pattern for}%
\typeout{** the default language instead.}%
\else
\language=\csname l@#1\endcsname
\fi
#2}}
\providecommand{\BIBdecl}{\relax}
\BIBdecl

\bibitem{DonelanEtAL}
J.~M. Donelan, R.~Kram, and A.~D. Kuo, ``Mechanical work for step-to-step
  transitions is a major determinant of the metabolic cost of human walking,''
  \emph{Journal of Experimental Biology}, vol. 205, no.~23, pp. 3717--3727,
  Dec. 2002.

\bibitem{KuoEtAL2005}
A.~D. Kuo, J.~M. Donelan, and A.~Ruina, ``Energetic consequences of walking
  like an inverted pendulum: Step-to-step transitions,'' \emph{Exercise and
  Sport Sciences Reviews}, vol.~33, no.~2, pp. 88--97, Apr. 2005.

\bibitem{AdolphEtAL2012}
K.~E. Adolph, W.~G. Cole, M.~Komati, J.~S. Garciaguirre, D.~Badaly, J.~M.
  Lingeman, G.~L.~Y. Chan, and R.~B. Sotsky, ``How do you learn to walk?
  {Thousands} of steps and dozens of falls per day,'' \emph{Psychological
  Science}, vol.~23, no.~11, pp. 1387--1394, Nov. 2012.

\bibitem{AdolphEtAL2003}
K.~E. Adolph, B.~Vereijken, and P.~E. Shrout, ``What changes in infant walking
  and why,'' \emph{Child Development}, vol.~74, no.~2, pp. 475--497, Mar. 2003.

\bibitem{Thelen1995}
E.~Thelen, ``Motor development: A new synthesis,'' \emph{American
  Psychologist}, vol.~50, no.~2, pp. 79--95, Feb. 1995.

\bibitem{SuttonEtAL1998}
R.~S. Sutton and A.~G. Barto, \emph{Reinforcement Learning: An Introduction},
  ser. Adaptive Computation and Machine Learning.\hskip 1em plus 0.5em minus
  0.4em\relax Cambridge, MA, USA: MIT Press, 1998.

\bibitem{RadosavovicEtAL2024}
I.~Radosavovic, T.~Xiao, B.~Zhang, T.~Darrell, J.~Malik, and K.~Sreenath,
  ``Real-world humanoid locomotion with reinforcement learning,'' \emph{Science
  Robotics}, vol.~9, no.~89, p. eadi9579, Apr. 2024.

\bibitem{GuEtAL2024}
X.~Gu, Y.-J. Wang, X.~Zhu, C.~Shi, Y.~Guo, Y.~Liu, and J.~Chen, ``Advancing
  humanoid locomotion: Mastering challenging terrains with denoising world
  model learning,'' in \emph{Robotics: Science and Systems ({RSS}) {XX}},
  Delft, Netherlands, Jul. 2024.

\bibitem{LiEtAL2025a}
Z.~Li, X.~B. Peng, P.~Abbeel, S.~Levine, G.~Berseth, and K.~Sreenath,
  ``Reinforcement learning for versatile, dynamic, and robust bipedal
  locomotion control,'' \emph{The International Journal of Robotics Research},
  vol.~44, no.~5, pp. 840--888, Apr. 2025.

\bibitem{RudinEtAL}
N.~Rudin, D.~Hoeller, P.~Reist, and M.~Hutter, ``Learning to walk in minutes
  using massively parallel deep reinforcement learning,'' in \emph{Proceedings
  of the 5th Conference on Robot Learning ({CoRL})}, ser. Proceedings of
  Machine Learning Research, vol. 164.\hskip 1em plus 0.5em minus 0.4em\relax
  PMLR, 2022, pp. 91--100.

\bibitem{FuEtAL2021}
Z.~Fu, A.~Kumar, J.~Malik, and D.~Pathak, ``Minimizing energy consumption leads
  to the emergence of gaits in legged robots,'' in \emph{Proceedings of the 5th
  Conference on Robot Learning ({CoRL})}, ser. Proceedings of Machine Learning
  Research, vol. 164.\hskip 1em plus 0.5em minus 0.4em\relax PMLR, 2022, pp.
  928--937.

\bibitem{JeonEtAL2023}
S.~H. Jeon, S.~Heim, C.~Khazoom, and S.~Kim, ``Benchmarking potential based
  rewards for learning humanoid locomotion,'' in \emph{{IEEE} International
  Conference on Robotics and Automation ({ICRA})}, London, UK, May 2023, pp.
  9204--9210.

\bibitem{KimEtAL2024}
Y.~Kim, H.~Oh, J.~Lee, J.~Choi, G.~Ji, M.~Jung, D.~Youm, and J.~Hwangbo, ``Not
  only rewards but also constraints: Applications on legged robot locomotion,''
  \emph{IEEE Transactions on Robotics}, vol.~40, pp. 2984--3003, 2024.

\bibitem{HuangEtAL2026}
W.~Huang, J.~Zhang, J.~Li, S.~Zhang, J.~Wu, J.~Wang, H.~Liu, Y.~Yang, and
  Y.~Su, ``{ECO}: Energy-constrained optimization with reinforcement learning
  for humanoid walking,'' \emph{IEEE Transactions on Automation Science and
  Engineering}, vol.~23, pp. 4861--4876, 2026.

\bibitem{PengEtAL2018}
X.~B. Peng, P.~Abbeel, S.~Levine, and M.~van~de Panne, ``{DeepMimic}:
  Example-guided deep reinforcement learning of physics-based character
  skills,'' \emph{ACM Transactions on Graphics}, vol.~37, no.~4, pp. 1--14,
  Aug. 2018.

\bibitem{PengEtAL2021}
X.~B. Peng, Z.~Ma, P.~Abbeel, S.~Levine, and A.~Kanazawa, ``{AMP}: Adversarial
  motion priors for stylized physics-based character control,'' \emph{ACM
  Transactions on Graphics}, vol.~40, no.~4, pp. 1--20, Aug. 2021.

\bibitem{MahmoodEtAL2019}
N.~Mahmood, N.~Ghorbani, N.~F. Troje, G.~Pons-Moll, and M.~J. Black, ``{AMASS}:
  Archive of motion capture as surface shapes,'' in \emph{{IEEE/CVF}
  International Conference on Computer Vision ({ICCV})}, Seoul, Korea, Oct.
  2019, pp. 5441--5450.

\bibitem{FarrisEtAL2012}
D.~J. Farris and G.~S. Sawicki, ``The mechanics and energetics of human walking
  and running: A joint level perspective,'' \emph{Journal of the Royal Society
  Interface}, vol.~9, no.~66, pp. 110--118, Jan. 2012.

\bibitem{McGeer1990a}
T.~McGeer, ``Passive dynamic walking,'' \emph{The International Journal of
  Robotics Research}, vol.~9, no.~2, pp. 62--82, Apr. 1990.

\bibitem{McGeer1990}
------, ``Passive walking with knees,'' in \emph{Proceedings of the {IEEE}
  International Conference on Robotics and Automation ({ICRA})}, Cincinnati,
  OH, USA, 1990, pp. 1640--1645.

\bibitem{HobbelenEtAL2007}
D.~G.~E. Hobbelen and M.~Wisse, ``A disturbance rejection measure for limit
  cycle walkers: The gait sensitivity norm,'' \emph{IEEE Transactions on
  Robotics}, vol.~23, no.~6, pp. 1213--1224, Dec. 2007.

\bibitem{CollinsEtAL2005}
S.~Collins, A.~Ruina, R.~Tedrake, and M.~Wisse, ``Efficient bipedal robots
  based on passive-dynamic walkers,'' \emph{Science}, vol. 307, no. 5712, pp.
  1082--1085, Feb. 2005.

\bibitem{ShiEtAL2023}
F.~Shi, Y.~Kojio, T.~Makabe, T.~Anzai, K.~Kojima, K.~Okada, and M.~Inaba,
  ``Reference-free learning bipedal motor skills via assistive force
  curricula,'' in \emph{Robotics Research ({ISRR} 2022)}, ser. Springer
  Proceedings in Advanced Robotics, A.~Billard, T.~Asfour, and O.~Khatib,
  Eds.\hskip 1em plus 0.5em minus 0.4em\relax Cham, Switzerland: Springer
  Nature Switzerland, 2023, vol.~27, pp. 304--320.

\bibitem{CaoEtAL2025}
Z.~Cao, Y.~Zhang, B.~Nie, H.~Lin, H.~Li, and Y.~Gao, ``Learning motion skills
  with adaptive assistive curriculum force in humanoid robots,''
  arXiv:2506.23125 [cs.RO], 2025.

\bibitem{KajitaEtAL2003}
S.~Kajita, F.~Kanehiro, K.~Kaneko, K.~Fujiwara, K.~Harada, K.~Yokoi, and
  H.~Hirukawa, ``Biped walking pattern generation by using preview control of
  zero-moment point,'' in \emph{{IEEE} International Conference on Robotics and
  Automation ({ICRA})}, Taipei, Taiwan, Sep. 2003, pp. 1620--1626.

\bibitem{GriffinEtAL2017}
R.~J. Griffin, G.~Wiedebach, S.~Bertrand, A.~Leonessa, and J.~Pratt, ``Walking
  stabilization using step timing and location adjustment on the humanoid
  robot, {Atlas},'' in \emph{{IEEE/RSJ} International Conference on Intelligent
  Robots and Systems ({IROS})}, Vancouver, BC, Canada, Sep. 2017, pp. 667--673.

\bibitem{WesterveltEtAL2007}
E.~R. Westervelt, J.~W. Grizzle, C.~Chevallereau, J.~H. Choi, and B.~Morris,
  \emph{Feedback Control of Dynamic Bipedal Robot Locomotion}, ser. Control and
  Automation.\hskip 1em plus 0.5em minus 0.4em\relax Boca Raton, FL, USA: CRC
  Press, Taylor \& Francis Group, 2007.

\bibitem{DongEtAL2011}
H.~Dong, M.~Zhao, and N.~Zhang, ``High-speed and energy-efficient biped
  locomotion based on {Virtual Slope Walking},'' \emph{Autonomous Robots},
  vol.~30, no.~2, pp. 199--216, Feb. 2011.

\bibitem{SiekmannEtAL2021}
J.~Siekmann, K.~Green, J.~Warila, A.~Fern, and J.~Hurst, ``Blind bipedal stair
  traversal via sim-to-real reinforcement learning,'' in \emph{Robotics:
  Science and Systems ({RSS}) {XVII}}, Virtual, Jul. 2021.

\bibitem{EscontrelaEtAL2022}
A.~Escontrela, X.~B. Peng, W.~Yu, T.~Zhang, A.~Iscen, K.~Goldberg, and
  P.~Abbeel, ``Adversarial motion priors make good substitutes for complex
  reward functions,'' in \emph{2022 {IEEE/RSJ} International Conference on
  Intelligent Robots and Systems ({IROS})}, 2022, pp. 25--32.

\bibitem{TangEtAL2024}
A.~Tang, T.~Hiraoka, N.~Hiraoka, F.~Shi, K.~Kawaharazuka, K.~Kojima, K.~Okada,
  and M.~Inaba, ``{HumanMimic}: Learning natural locomotion and transitions for
  humanoid robot via {Wasserstein} adversarial imitation,'' in \emph{2024
  {IEEE} International Conference on Robotics and Automation ({ICRA})}, 2024.

\bibitem{VollenweiderEtAL2023}
E.~Vollenweider, M.~Bjelonic, V.~Klemm, N.~Rudin, J.~Lee, and M.~Hutter,
  ``Advanced skills through multiple adversarial motion priors in reinforcement
  learning,'' in \emph{2023 {IEEE} International Conference on Robotics and
  Automation ({ICRA})}, 2023, pp. 5120--5126.

\bibitem{ChenEtAL2025b}
Z.~Chen, X.~He, Y.-J. Wang, Q.~Liao, Y.~Ze, Z.~Li, S.~S. Sastry, J.~Wu,
  K.~Sreenath, S.~Gupta, and X.~B. Peng, ``Learning smooth humanoid locomotion
  through {Lipschitz}-constrained policies,'' in \emph{{IEEE/RSJ} International
  Conference on Intelligent Robots and Systems ({IROS})}, Hangzhou, China, Oct.
  2025, pp. 4743--4750.

\bibitem{XiaEtAL2025}
B.~Xia, B.~Li, J.~Lee, M.~Scutari, and B.~Chen, ``The {Duke Humanoid}: Design
  and control for energy-efficient bipedal locomotion using passive dynamics,''
  in \emph{{IEEE/RSJ} International Conference on Intelligent Robots and
  Systems ({IROS})}, Hangzhou, China, Oct. 2025, pp. 6579--6586.

\bibitem{DonelanEtAL2002}
J.~M. Donelan, R.~Kram, and A.~D. Kuo, ``Simultaneous positive and negative
  external mechanical work in human walking,'' \emph{Journal of Biomechanics},
  vol.~35, no.~1, pp. 117--124, Jan. 2002.

\bibitem{WisseEtAL2007}
M.~Wisse, D.~G.~E. Hobbelen, and A.~L. Schwab, ``Adding an upper body to
  passive dynamic walking robots by means of a bisecting hip mechanism,''
  \emph{IEEE Transactions on Robotics}, vol.~23, no.~1, pp. 112--123, Feb.
  2007.

\bibitem{GarciaEtAL1998}
M.~Garcia, A.~Chatterjee, A.~Ruina, and M.~Coleman, ``The simplest walking
  model: Stability, complexity, and scaling,'' \emph{Journal of Biomechanical
  Engineering}, vol. 120, no.~2, pp. 281--288, Apr. 1998.

\bibitem{BaubyEtAL2000}
C.~E. Bauby and A.~D. Kuo, ``Active control of lateral balance in human
  walking,'' \emph{Journal of Biomechanics}, vol.~33, no.~11, pp. 1433--1440,
  Nov. 2000.

\bibitem{HandfordEtAL2014}
M.~L. Handford and M.~Srinivasan, ``Sideways walking: Preferred is slow, slow
  is optimal, and optimal is expensive,'' \emph{Biology Letters}, vol.~10,
  no.~1, p. 20131006, Jan. 2014.

\bibitem{FlorensaEtAL2018}
C.~Florensa, D.~Held, X.~Geng, and P.~Abbeel, ``Automatic goal generation for
  reinforcement learning agents,'' in \emph{Proceedings of the 35th
  International Conference on Machine Learning ({ICML})}, ser. Proceedings of
  Machine Learning Research, vol.~80, 2018, pp. 1515--1528.

\bibitem{NarvekarEtAL2020}
S.~Narvekar, B.~Peng, M.~Leonetti, J.~Sinapov, M.~E. Taylor, and P.~Stone,
  ``Curriculum learning for reinforcement learning domains: A framework and
  survey,'' \emph{Journal of Machine Learning Research}, vol.~21, no. 181, pp.
  1--50, 2020.

\bibitem{FukunagaEtAL2001}
T.~Fukunaga, K.~Kubo, Y.~Kawakami, S.~Fukashiro, H.~Kanehisa, and C.~N.
  Maganaris, ``In vivo behaviour of human muscle tendon during walking,''
  \emph{Proceedings of the Royal Society B: Biological Sciences}, vol. 268, no.
  1464, pp. 229--233, Feb. 2001.

\bibitem{CollinsEtAL2015}
S.~H. Collins, M.~B. Wiggin, and G.~S. Sawicki, ``Reducing the energy cost of
  human walking using an unpowered exoskeleton,'' \emph{Nature}, vol. 522, no.
  7555, pp. 212--215, Jun. 2015.

\bibitem{WinterEtAL1989}
D.~A. Winter, N.~Pluck, and J.~F. Yang, ``Backward walking: A simple reversal
  of forward walking?'' \emph{Journal of Motor Behavior}, vol.~21, no.~3, pp.
  291--305, Sep. 1989.

\bibitem{GrassoEtAL1998}
R.~Grasso, L.~Bianchi, and F.~Lacquaniti, ``Motor patterns for human gait:
  Backward versus forward locomotion,'' \emph{Journal of Neurophysiology},
  vol.~80, no.~4, pp. 1868--1885, Oct. 1998.

\end{thebibliography}

\end{document}